\documentclass{article}

\usepackage{arxiv}

\input{paper_style.sty}
\usepackage{natbib}
\usepackage{moreverb,url}
\usepackage[colorlinks,bookmarksopen,bookmarksnumbered,citecolor=red,urlcolor=red]{hyperref}
\usepackage{booktabs}

\title{Clarke Transform -- A Fundamental Tool for Continuum Robotics}

\author{
    \normalfont Reinhard Grassmann$^{1}$
    \and
    \normalfont Anastasiia Senyk$^{1,2}$
    \and
    \normalfont Jessica Burgner-Kahrs$^{1}$ 
    \thanks{\hspace*{-1.8em}
    $^1$Continuum Robotics Laboratory, Department of Mathematical \&
    Computational Sciences, University of Toronto, Mississauga, ON,
    Canada \newline
    $^2$Faculty of Applied Science, Ukrainian Catholic University, Lviv, 79026, Ukraine \newline
    Corresponding author: Reinhard M. Grassmann, e-mail: \texttt{reinhard.grassmann@utoronto.ca}}
}

\date{}

\renewcommand\footnotemark{}

\begin{document}

\maketitle

\begin{abstract}
This article introduces the Clarke transform and Clarke coordinates, which present a solution to the disengagement of an arbitrary number of coupled displacement actuation of continuum and soft robots.
The Clarke transform utilizes the generalized Clarke transformation and its inverse to reduce any number of joint values to a two-dimensional space without sacrificing any significant information.
This space is the manifold of the joint space and is described by two orthogonal Clarke coordinates.
Application to kinematics, sampling, and control are presented.
By deriving the solution to the previously unknown forward robot-dependent mapping for an arbitrary number of joints, the forward and inverse kinematics formulations are branchless, closed-form, and singular-free.
Sampling is used as a proxy for gauging the performance implications for various methods and frameworks, leading to a branchless, closed-form, and vectorizable sampling method with a 100 percent success rate and the possibility to shape desired distributions.
Due to the utilization of the manifold, the fairly simple constraint-informed, two-dimensional, and linear controller always provides feasible control outputs.
On top of that, the relations to improved representations in continuum and soft robotics are established, where the Clarke coordinates are their generalizations.

The Clarke transform offers valuable geometric insights and paves the way for developing approaches directly on the two-dimensional manifold within the high-dimensional joint space, ensuring compliance with the constraint.
While being an easy-to-construct linear map, the proposed Clarke transform is mathematically consistent, physically meaningful, as well as interpretable and contributes to the unification of frameworks across continuum and soft robots.
\end{abstract}

\maketitle

\section{Introduction}
\label{sec:intro}

Robots are described in the joint space, while the interaction with the environment and humans is described in the task space.  
To bridge the joint space and the task space, an intermediate space is used to take advantage of the robot's specific geometry.
For serial-kinematic robots, line coordinates are used to construct the intermediate space.
The most prominent line coordinates are described by Denavit-Hartenberg parameters, where \cite{Paul_book_1981} and \cite{Craig_book_2005} popularize the standard form and the modified form, respectively.
For continuum robots, geometry descriptions related to circular arcs are used.
The most prominent arc representation popularized by \cite{WebsterJones_IJRR_2010} assumes constant curvature.
The intermediate space encompassing an arc representation is called arc space.
Note that this intermediate pace is often called configuration space \citep{WebsterJones_IJRR_2010, RaoBurgner-Kahrs_et_al_Frontiers_2021} and different representations exist, see Table~\ref{tab:arc_space_representation} and Appendix~\ref{appendix:arc_space_representation}.

\begin{table*}[!t]
    \renewcommand*{\arraystretch}{1.4}
    \caption{
        Various improved state representations and their connection to the joint space.
        All authors use actuation length $l_i$ as joint space representation.
        Notations are adapted to our notation, where the preferred variable name for the improved state representation has been retained, and the distance between the imagery center-line and actuation location is denoted by $d$. 
        Furthermore, a notation indicating the use of piecewise constant curvature is omitted by dropping an additional index.
    }
    \label{tab:overview_improved_state_representation}
    \centering
    \begin{tabular}{r rrr} 
        \toprule
        \multicolumn{1}{N}{Reference} & 
        \multicolumn{1}{N}{Parameterization} & 
        \multicolumn{1}{N}{number of joints} & 
        \multicolumn{1}{N}{actuation type} \\
		\cmidrule(r){1-1}
		\cmidrule(lr){2-2}
		\cmidrule(lr){3-3}
		\cmidrule(l){4-4}
        \\[-.875em]
        \cite{DellaSantinaBicchiRus_RAL_2020} & $\Delta_{x} = \dfrac{l_{3} - l_{1}}{2} \quad\text{and}\quad \Delta_{y} = \dfrac{l_{4} - l_{2}}{2}$ & $4$ & pneumatic \\[.75em]
        \cite{AllenAlbert_et_al_RoboSoft_2020} & $u = \dfrac{l_2 - l_3}{\sqrt{3}d} \quad\text{and}\quad v = \dfrac{\left(l_1 + l_2 + l_3\right)/3 - l_1}{d}$ & $3$ & actuation length \\[.75em]
        \cite{AllenAlbert_et_al_RoboSoft_2020} & $u = \dfrac{l_2 - l_4}{d} \quad\text{and}\quad v = \dfrac{l_3 - l_1}{d}$ & $4$ & actuation length \\[.75em]
        \cite{DianGuo_et_al_Access_2022} & $\Delta x = \dfrac{l_2 + l_3 - 2l_1}{3} \quad\text{and}\quad \Delta y = \dfrac{l_3 - l_2}{\sqrt{3}}$ & $3$ & cable-driven \\[.75em]
		\cmidrule(r){1-1}
		\cmidrule(lr){2-2}
		\cmidrule(lr){3-3}
		\cmidrule(l){4-4}
        \multicolumn{1}{c}{\textit{gap in the literature}} & \multicolumn{1}{c}{\textit{unknown yet}} & $n$ & displacement \\[.25em]
        \bottomrule
    \end{tabular}
\end{table*}

Coordinate singularities can occur that are not a singularity of the robot but rather an artifact of the parameterization of the representation used.
Especially, the curvature-angle and angle-angle representations are prone to have a coordinate singularity, where a straight configuration causes the curvature to approach zero and the bending radius to approach infinity, resulting in an indeterminate form in the kinematics formulation.
This problem is not limited to kinematics and affects velocity kinematics \citep{JonesWalker_ICRA_2007}, dynamics \citep{TatliciogluWalkerDawson_IROS_2007, RoneBen-Tzvi_TRO_2013, RendaSeneviratne_et_al_TRO_2018}, and model-based control \citep{DellaSantinaBicchiRus_RAL_2020}, among other things.

\begin{figure}[b]
\centering
    \includegraphics[width=0.45\columnwidth]{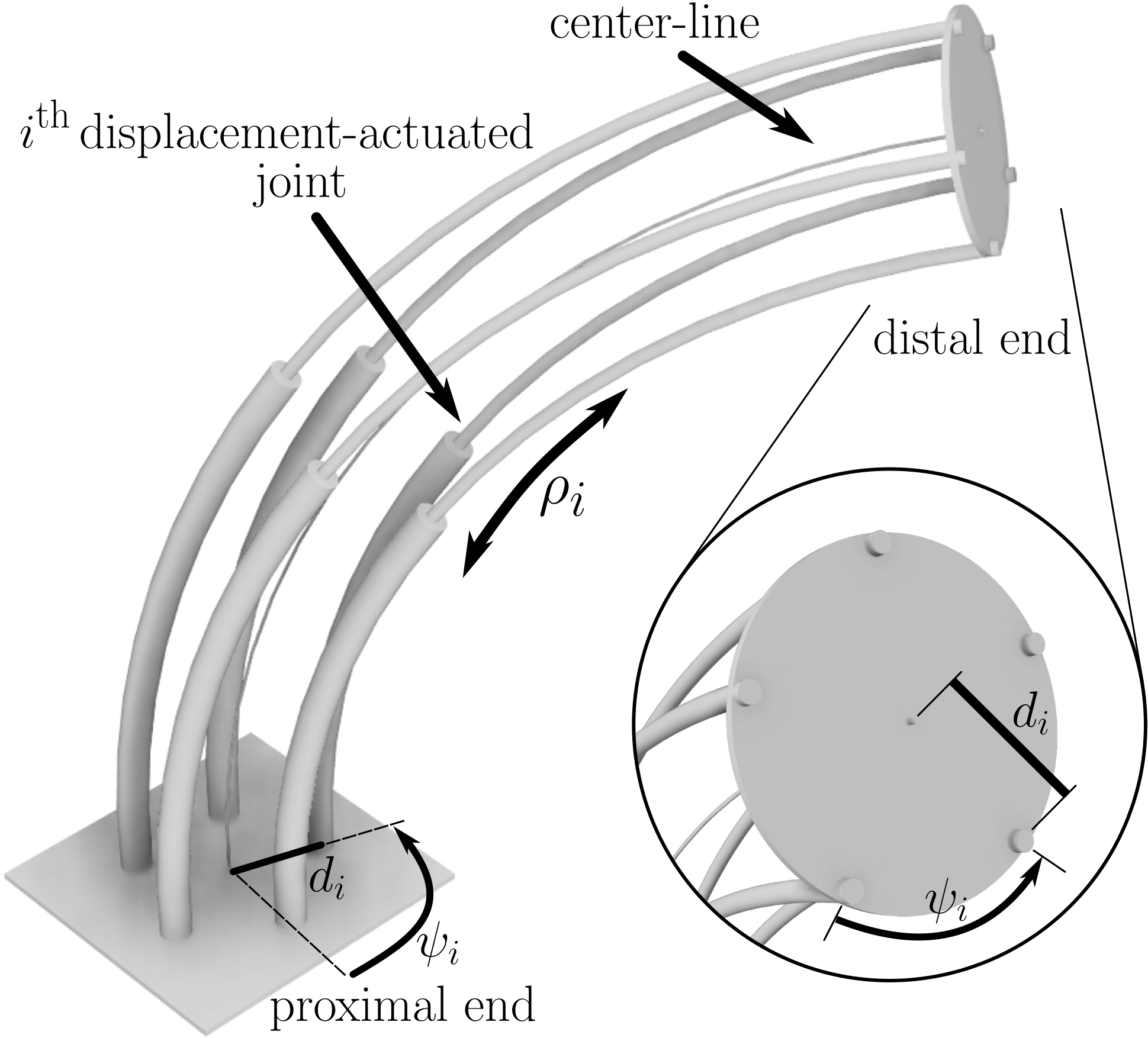}
    \caption{
    Schematics of a displacement-actuated robot with sufficient smooth center-line with fixed length $l$.
    Each of the displacement-actuated joints is equally distributed, described by $d_i$ and $\psi_i$.
    Its value is described by a displacement $\rho_i$.
    }
    \label{fig:displacement_actuated_robot}
\end{figure}

To overcome the coordinate singularity, several authors propose improved state representations \citep{QuLau_et_al_ROBIO_2016, DellaSantinaBicchiRus_RAL_2020, AllenAlbert_et_al_RoboSoft_2020, CaoXie_et_al_JMR_2022, DianGuo_et_al_Access_2022} for their respective robot type, see Table~\ref{tab:overview_improved_state_representation}.
By closer inspection of the kinematic description, one can see that those robot types are described similarly.
\cite{AllenAlbert_et_al_RoboSoft_2020} make a similar observation, where the joint, \textit{e.g.}, artificial muscles or tendon length, can be described as actuation length.
However, one can argue that relative displacement should be preferred over absolute actuation length, see Appendix~\ref{appendix:absolute_actuation_length}.
Therefore, a large class of soft robots and continuum robots can be classified as displacement-actuated robots, where the actuation occurs symmetrically and equidistantly \textit{w.r.t.} an imaginary center-line, see Figure~\ref{fig:displacement_actuated_robot} for visual aid.
For those types of spatial robots with a single segment, the workspace is a \SI{2}{dof} dome in task space regardless of the number of displacement joints actuating the segment, see Appendix~\ref{appendix:observation} for an intuitive example.
That the underlying problem is \SI{2}{dof} is also reflected in the dimensionality of arc parameters and improved state representations hinting at a \SI{2}{dof} manifold embedded in the joint space.

In this article, we examine the commonly used displacement-joint constraint, \textit{i.e.}, the sum of all displacements should be zero,
\begin{align}
	\sum_{i=1}^{n} \rho_{i} = 0, 
    \label{eq:sum_rho}
\end{align}
and build on our intuition (see Appendix~\ref{appendix:intuition}) to derive the Clarke transform and Clarke coordinates.
For that, we propose a displacement representation and identify the \SI{2}{dof} manifold by utilizing a generalized Clarke transformation matrix.
To illustrate the practicability, application to kinematic modelling, sampling, and joint space control are presented.
While we assume constant curvature for some applications, the Clarke transform and Clarke coordinates are not limited to this assumption.
In particular, the contribution of this article includes:
\begin{itemize}
    \item Development and proposal of the Clarke transform to map between the $n\,\SI{}{dof}$ joint space and the \SI{2}{dof} manifold (Sec.~\ref{sec:clarke_transform})
    \item Introduction of displacement representation and Clarke coordinates that unify and generalize improved state representations (Sec.~\ref{sec:clarke_transform})
    \item Identification of \SI{2}{dof} manifold embedded in the joint space (Sec.~\ref{sec:clarke_transform})
    \item Formulation of compact and closed-form expressions to the robot-dependent mappings that extend known solutions to an arbitrary number of joints (Sec.~\ref{sec:kinematics})
    \item Presentation of a generalized, linear, compact, and closed-form solutions to the kinematics (Sec.~\ref{sec:kinematics}), sampling (Sec.~\ref{sec:sampling}), and joint control (Sec.~\ref{sec:control})
    \item Explanation of geometric insights on the displacement constraint and coordinate singularities
\end{itemize}

\section{Clarke Transform}
\label{sec:clarke_transform}

This section starts by introducing a novel representation of displacement-actuated joints.
Afterwards, we introduce the generalized Clarke transformation matrix \textit{w.r.t.} number of displacement $n \geq 3$ and apply it to map between joint space and \SI{2}{dof} manifold.
The generalized Clarke transformation matrix is the base for the Clarke transform and Clarke coordinates, which are introduced at the end of this section.
Note that, in prior work \citep{GrassmannBurgner-Kahrs_ICRA_EA_2024}, we briefly introduced the concept of the Clarke transform and its potential application to continuum robotics.

\subsection{Displacement Representation and Embedded Manifold}

Before defining a set for the \SI{2}{dof} manifold embedded in the $n\,\SI{}{dof}$ joint space for $n$ displacements, we make the following observation.
For $n$ equally distributed displacement-actuated joints, the angle and radius
\begin{align}
    \psi_i = \dfrac{2\pi}{n} (i - 1)
    \quad\text{and}\quad
    d_i = d
    ,
    \label{eq:psi}
\end{align}
respectively, describe the location of the $i\textsuperscript{th}$ joint in the cross-section in terms of polar coordinates, see Figure~\ref{fig:displacement_actuated_robot}.
Using the angle in \eqref{eq:psi}, we can find the following trigonometric sums 
\begin{align}
	\sum_{i=1}^{n} \sin\left(\psi_i\right) &= 0 \quad\text{and} \label{eq:sum_sine}\\
	\sum_{i=1}^{n} \cos\left(\psi_i\right) &= 0. \label{eq:sum_cosine}
\end{align}
Since each of the sums is zero, the linear combination of both sums is zero as well.
This implies the coefficients of this combination, denoted as $\rho_\text{Re}$ and $\rho_\text{Im}$, act as free parameters. 
This leads to our preliminary check given by
\begin{align}
	\rho_\text{Re}\sum_{i=1}^{n} \cos\left(\psi_i\right) + 
	\rho_\text{Im}\sum_{i=1}^{n} \sin\left(\psi_i\right) = 0.
    \label{eq:sanity_check}
\end{align}

\begin{proof}
    Its proof involves the proof of \eqref{eq:sum_sine} and \eqref{eq:sum_cosine} being straightforward when using the Lagrange's trigonometric identities for  $\sum_{j=0}^{n} \sin\left(\alpha j\right)$ and $\sum_{j=0}^{n} \cos\left(\alpha j\right)$.
    First, a change of index is performed, \textit{i.e.}, $j = i - 1$, such that \eqref{eq:psi} simplifies to $(2\pi/n)j = \alpha j$ and the summations \eqref{eq:sum_sine} and \eqref{eq:sum_cosine} are defined from $j = 0$ to $n-1$.
    Second, note that the upper bound of summation is shifted; therefore, the summand for $j=n$ needs to be added to get Lagrange's trigonometric identity.
    Afterward, apply the Lagrange's trigonometric identities to verify \eqref{eq:sum_sine} and \eqref{eq:sum_cosine}.
    Finally, the linear combination of \eqref{eq:sum_sine} and \eqref{eq:sum_cosine} is \eqref{eq:sanity_check}, which completes the proof.
\end{proof}

Based on this observation, we define a representation for displacement-actuated joints in its \textit{rectangular form} given by
\begin{align}
    \rho_i = \rho_\text{Re}\cos{\psi_i} + \rho_\text{Im}\sin{\psi_i}.
    \label{eq:rho_rectangular_form}
\end{align}
It has the same two variables for all $i$.

\begin{proof}
    Substitute \eqref{eq:rho_rectangular_form} into \eqref{eq:sum_rho} and identify \eqref{eq:sanity_check}.
\end{proof}

Note that \eqref{eq:rho_rectangular_form} is a linear combination of trigonometric functions.
Using harmonic addition, sine and cosine functions can be linearly combined.
In general, the result is a shifted cosine function with scaled amplitude, \textit{i.e.},
\begin{align}
    D_i\cos(\psi_i - \theta) = A_i\cos(\psi_i) + B_i\sin(\psi_i),
    \label{eq:harmonic_addition}
\end{align}
where polar coordinates $D_i$ and $\theta$ are defined as
\begin{align}
    D_i = \sqrt{A_i^2 + B_i^2} \quad\text{and}\quad \theta = \arctantwo\left({B_i, A_i}\right),
    \label{eq:harmonic_addition_D_theta}
\end{align}
respectively.
The right-hand side of \eqref{eq:harmonic_addition} is the rectangular form and can be obtained by expanding the left-hand side of \eqref{eq:harmonic_addition} called the polar form.
Using a scaled trigonometric addition formula, \textit{i.e.}, $D_i\cos(\psi_i - \theta) = D_i\cos(\psi_i)\cos(\theta) + D_i\sin(\psi_i)\sin(\theta)$, the rectangular coordinates
\begin{align}
    A_i = D_i\cos(\theta) \quad\text{and}\quad B_i = D_i\sin(\theta)
    \label{eq:harmonic_addition_A_B}
\end{align}
can be defined.
The harmonic addition allows us to define
\begin{align}
    \rho_i = \sqrt{\rho_\text{Re}^2 + \rho_\text{Im}^2}\cos{\left(\psi_i - \arctantwo\left({\rho_\text{Im},\ \rho_\text{Re}}\right)\right)}
    \label{eq:rho_polar_form}
\end{align}
as the corresponding \textit{polar form} of \eqref{eq:rho_rectangular_form}.

Using $n+2$ variables including $\rho_\text{Re}$ and $\rho_\text{Im}$, we get $n$ equations for $n$ displacements defined by \eqref{eq:rho_rectangular_form} or \eqref{eq:rho_polar_form}.
For convenience and compactness, we use \eqref{eq:rho_rectangular_form} to define the set $\mathbf{Q}$ as an expression of the \SI{2}{dof} manifold embedded in the $n\,\SI{}{dof}$ joint space for $n$ displacement joints as
\begin{align}
    \mathbf{Q}=
    \bigl\{\bigr.\,&\left(\rho_1, \cdots, \rho_n\right) \in \mathbb{R}^n \,\, \bigl|\bigr. \,\, \forall i \in \left[1, n\right] \subset \mathbb{N}:\nonumber\\
    \,&\rho_i = \rho_\text{Re}\cos{\psi_i} + \rho_\text{Im}\sin{\psi_i}\,\,\wedge\nonumber\\
    \,&\left(\rho_\text{Re}, \rho_\text{Im}\right) \in \mathbb{R}^2\,\bigl.\bigr\},
    \label{eq:manifold_n_tendons}
\end{align}
where $\rho_i$ is defined for $i = \{1, \dots, n \}$ as well as $\rho_\text{Re} \in \mathbb{R}$ and $\rho_\text{Im} \in \mathbb{R}$.
The local coordinates of the manifold are $\rho_\text{Re}$ and $\rho_\text{Im}$.

\subsection{Clarke Transformation Matrix}
\label{sec:Clarke_transformation_matrix}

The Clarke transformation matrix reduces three electric current values to two values, whereas the third value is zero.
The corresponding matrix is given here in its generic form,
\begin{align}
	\boldsymbol{M}_\text{Clarke}
    = k_0
    \begin{bmatrix} 
        1 & -\dfrac{1}{2} & -\dfrac{1}{2} \\[1em]
        0 & \dfrac{\sqrt{3}}{2} & -\dfrac{\sqrt{3}}{2} \\[1em]
        k_1 & k_1 & k_1 \\
    \end{bmatrix},
    \label{eq:M_Clarke}
\end{align}
where $k_0$ and $k_1$ are free parameters.
Two main variances of \eqref{eq:M_Clarke} are usually considered: power-invariant and amplitude-invariant forms.
The former has parameters set to $k_0 = \sqrt{2/3}$ and $k_1 = \sqrt{2}/2$.
The latter is also called the standard form with $k_0 = 2/3$ and $k_1 = 1/2$.
The inverse Clarke transformation allows the recovery of the original three values without information loss. 
Note that \eqref{eq:M_Clarke} is sometimes given as its inverse for a given set of parameters, \textit{e.g.}, \citep{Willems_TOE_1969, EboulePretoriusMbuli_APPEEC_2019}.

\subsection{Generalized Clarke Transformation Matrix}
\label{sec:disentaglement_ClarkeTransform}

Looking at the reduced form of \eqref{eq:M_Clarke}, \textit{i.e.}, omitting the last row, we find the pattern for $n = 3$ in \eqref{eq:psi}.
Neglecting the coefficient $k_0$ in \eqref{eq:M_Clarke} for now, the first row is $\left[\cos\left(\psi_1\right), \cos\left(\psi_2\right), \cos\left(\psi_3\right)\right]$ and the second row is $\left[\sin\left(\psi_1\right), \sin\left(\psi_2\right), \sin\left(\psi_3\right)\right]$.
By following the pattern, we can generalize Clarke transformation matrix \textit{w.r.t.} the number of variables to be transformed, \textit{i.e.}, $n$. 
By multiplying the representation \eqref{eq:rho_rectangular_form} with a potential generalized Clarke transformation matrix with the pattern described above, sums of squared and mixed trigonometric functions occur, \textit{i.e.}, $\sin^2\left(\psi_i\right)$, $\cos^2\left(\psi_i\right)$, and $\sin\left(\psi_i\right)\cos\left(\psi_i\right)$.

To deal with the three different sums of trigonometric functions, we establish three identities.
The first identity is 
\begin{align}
	\sum_{i=1}^{n} \sin\left(\psi_i\right)\cos\left(\psi_i\right) &= 0,
    \label{eq:sum_sinecosine}
\end{align}
which is related to \eqref{eq:sum_sine}.
\begin{proof}
    Rewrite \eqref{eq:sum_sinecosine} by using the double-angle formula, \textit{i.e.}, $\sin\left(2\psi_i\right) = 2\sin\left(\psi_i\right)\cos\left(\psi_i\right)$.
    Afterward, identify \eqref{eq:sum_sine}.
    Note that $\alpha$ for Lagrange's trigonometric identity needs to be constant \textit{w.r.t.} the variable of summation.
    For the proof of \eqref{eq:sanity_check}, $\alpha = 2\pi/n$, and, for this proof, $\alpha = 4\pi/n$.
\end{proof}
While the previous encounter identities are zero, see \eqref{eq:sum_sine}, \eqref{eq:sum_cosine}, and \eqref{eq:sum_sinecosine}, a sum of squared trigonometric functions is not zero.
This is shown by the identities given by 
\begin{align}
	\sum_{i=1}^{n} \sin^2\left(\psi_i\right) &= \dfrac{n}{2} \label{eq:sum_sinesine} \quad\text{and}\\
	\sum_{i=1}^{n} \cos^2\left(\psi_i\right) &= \dfrac{n}{2}. \label{eq:sum_cosinecosine}
\end{align}
\begin{proof}
    By using double-angle formulae for cosine, \textit{i.e.}, $\cos\left(2\psi_i\right) = 1 - 2\sin^2\left(\psi_i\right) = 2\cos^2\left(\psi_i\right) - 1$, the identities \eqref{eq:sum_sinesine} and \eqref{eq:sum_cosinecosine} can be decomposed into \eqref{eq:sum_cosine} and a sum of a constant, \textit{i.e.}, $1/2$.
    Consequently, both sums evaluate to $n/2$.
\end{proof}

Once we have recognized a pattern and presented the identities, we take the idea of using $\boldsymbol{M}_\text{Clarke}$, see Sec.~\ref{sec:Clarke_transformation_matrix}, and define
\begin{align}
    \begin{bmatrix} 
        \rho_\text{Re} \\ \rho_\text{Im} 
    \end{bmatrix}
    = 
    \boldsymbol{M}_\mathcal{P} 
    \begin{bmatrix} 
        \rho_1 & \rho_2 & \cdots & \rho_{n-1} & \rho_n 
    \end{bmatrix}\transpose
    \label{eq:forward_mapping}
\end{align}
with $\boldsymbol{M}_\mathcal{P}$ defined as
\begin{align}
	\boldsymbol{M}_\mathcal{P}
    =
    \dfrac{2}{n}\!
	\begin{bmatrix}
		\cos\left(0\right) & \cos\left(2\pi\dfrac{1}{n}\right) & \cdots & \cos\left(2\pi\dfrac{n-1}{n}\right)\\[1em]
		\sin\left(0\right) & \sin\left(2\pi\dfrac{1}{n}\right) & \cdots & \sin\left(2\pi\dfrac{n-1}{n}\right)
	\end{bmatrix}
    ,
	\label{eq:MP}
\end{align}
which is a generalized Clarke transformation matrix.
The generalized Clarke transformation matrix \eqref{eq:MP} relates the $n$ displacements defined by \eqref{eq:rho_rectangular_form} to the two variables, \textit{i.e.}, $\rho_\mathrm{Re}$ and $\rho_\mathrm{Im}$ of \eqref{eq:manifold_n_tendons}.
\begin{proof}
    Expanding the right-hand side of \eqref{eq:forward_mapping} gives a term with mixed trigonometric functions and squared trigonometric functions.
    While the mixed trigonometric functions sum up to zero, see \eqref{eq:sum_sinecosine}, the squared terms result in a constant value, \textit{i.e.}, \eqref{eq:sum_cosinecosine} for the first row and \eqref{eq:sum_sinesine} for the second row.
    Considering the scaling factor $2/n$ of \eqref{eq:MP} chosen to compensate for \eqref{eq:sum_sinesine} and \eqref{eq:sum_cosinecosine}, the right-hand side equals the left-hand side of \eqref{eq:forward_mapping}.
\end{proof}
Note that, by using \eqref{eq:rho_rectangular_form}, \eqref{eq:sum_sinecosine}, \eqref{eq:sum_sinesine}, and \eqref{eq:sum_cosinecosine}, the magnitude of $\left[\rho_1, \rho_2, \cdots, \rho_{n-1}, \rho_n \right]\transpose$ is $\sqrt{n/2}\sqrt{\rho_\text{Re}^2 + \rho_\text{Im}^2}$.
Hence, this representation scales with its length.
On the contrary, $\left[\rho_\text{Re}, \rho_\text{Im}\right]\transpose$ has a feature that its magnitude, \textit{i.e.}, $\sqrt{\rho_\text{Re}^2 + \rho_\text{Im}^2}$, does not change \textit{w.r.t.} $n$.
Therefore, \eqref{eq:forward_mapping} is amplitude invariant.

To recover $\rho_i$, the system of linear equations given by 
\begin{align}
    \begin{bmatrix} 
        \rho_1 & \rho_2 & \cdots & \rho_{n-1} & \rho_n 
    \end{bmatrix}\transpose
    = 
    \boldsymbol{M}_\mathcal{P}^{-1}
    \begin{bmatrix} 
        \rho_\text{Re} \\ \rho_\text{Im} 
    \end{bmatrix}
    \label{eq:inverse_mapping}
\end{align}
is solved, where the inverse mapping $\boldsymbol{M}_\mathcal{P}^{-1}$ is defined as
\begin{align}
	\boldsymbol{M}_\mathcal{P}^{-1}
    =
	\begin{bmatrix}
		\cos\left(0\right) & \sin\left(0\right) \\[1em]
        \cos\left(2\pi\dfrac{1}{n}\right) & \sin\left(2\pi\dfrac{1}{n}\right) \\[1em]
        \vdots & \vdots\\[1em]
        \cos\left(2\pi\dfrac{n-1}{n}\right) & \sin\left(2\pi\dfrac{n-1}{n}\right)
	\end{bmatrix}
    .
	\label{eq:MP_inverse}
\end{align}
From \eqref{eq:inverse_mapping} and \eqref{eq:MP_inverse}, one can see that the contributions of the location, \textit{i.e.}, $\cos\left(\psi_i\right)$ and $\sin\left(\psi_i\right)$, are added to $\rho_\text{Re}$ and $\rho_\text{Im}$.
Note that $\left(\right)^{-1}$ in \eqref{eq:MP_inverse} indicates the inverse transformation not a matrix inverse of $\boldsymbol{M}_\mathcal{P}$ stated in \eqref{eq:MP}.
In fact, \eqref{eq:MP} is a non-square matrix and \eqref{eq:MP_inverse} is the right-inverse of \eqref{eq:MP}.

\subsection{Common Choices of n}

The generalized Clarke transformation matrix \eqref{eq:MP} is the extension of \eqref{eq:M_Clarke}, where the scaling factor $k_0$ is $2/n$, and the last rows are omitted.
Note that each omitted row leads to a zero-sequence component meaning that the value is always zero.
To gain more intuition and as a sanity check, we confirm \eqref{eq:MP} for common choices of $n$, \textit{i.e.}, $n = \left\{3, 4\right\}$.

As expected, for $n = 3$,
\begin{align}
	\boldsymbol{M}_\mathcal{P}\left(n=3\right)
    = \frac{2}{3}
    \begin{bmatrix} 
        1 & -\dfrac{1}{2} & -\dfrac{1}{2} \\[1em]
        0 & \dfrac{\sqrt{3}}{2} & -\dfrac{\sqrt{3}}{2}
    \end{bmatrix}
    \nonumber
\end{align}
is the simplified matrix of \eqref{eq:M_Clarke}, where the last row is removed and $k_0 = 2/3$.
It is the amplitude-invariant form.
Furthermore, $\MPinv\left(n=3\right)$ results in the known inverse transformation matrix, \textit{cf.} \citep{ORourkeKirtley_et_al_TEC_2019}.

For a $n = 4$ configuration with two pairs of differential actuation, the transformation matrix
\begin{align}
	\boldsymbol{M}_\mathcal{P}\left(n=4\right)
    = \frac{1}{2}
	\begin{bmatrix}
		1 & 0 & -1 & 0\\
		0 & 1 & 0 & -1 
	\end{bmatrix}
    \nonumber
\end{align}
shows a familiar pattern.
The result of the transformation leads to $\rho_\text{Re} = \left(\rho_1 - \rho_3\right)/2$ and $\rho_\text{Im} = \left(\rho_2 - \rho_4\right)/2$.
Both constraints for $n = 4$, see Appendix~\ref{appendix:observation}, \textit{i.e.}, $\rho_1 + \rho_3 = 0$ and $\rho_2 + \rho_4 = 0$, reveals that $\rho_\text{Re}$ and $\rho_\text{Im}$ correspond to a different bending plane.
This indicates that $\rho_\text{Re}$ directly relates the curvature, \textit{i.e.}, $\kappa_x$ for a common convention, of a plane that is spanned by the base frame's $x$-axis and $z$-axis.
Similarly, $\rho_\text{Im}$ directly relates to $\kappa_y$.
Furthermore, the relation indicates that $\rho_\text{Re}$ and $\rho_\text{Im}$ are orthogonal to each other.

\subsection{Clarke Transform and Clarke Coordinates}

After having defined the generalized Clarke transformation matrix \eqref{eq:MP} and its inverse \eqref{eq:MP_inverse}, let us now introduce the Clarke transform. 
As mentioned before, \eqref{eq:MP} can be used to map the joint space quantities $\rho_i$ of a displacement-actuated robot onto the manifold.
The Clarke transform describes this process, where the forward transform converts quantities from the joint space $\mathbf{Q}$ to quantities in the manifold, and the inverse transform reverts quantities in the manifold to joint space quantities.
In this context, the term Clarke transform refers to an operation.

The local coordinates of the \SI{2}{dof} manifold embedded in the $n\,\SI{}{dof}$ joint space for $n$ actuator displacements $\rho_i$ that can be combined as vector, \textit{i.e.},
\begin{align}
    \rhovec
    = 
    \begin{bmatrix} 
        \rho_1 & \rho_2 & \cdots & \rho_{n-1} & \rho_n 
    \end{bmatrix}\transpose \in \mathbf{Q} \subset \mathbb{R}^n
    ,
\end{align}
are the two free parameters of $\rho_i$, \textit{i.e.}, $\rho_\mathrm{Re}$ and $\rho_\mathrm{Im}$.
The free parameters can be combined into a vector, \textit{i.e.},
\begin{align}
    \rhoclarke = \left[ \rhoreal, \rhoim \right]\transpose  \in \mathbb{R}^2.
    \label{eq:rho_clarke}
\end{align}
We call them \textit{Clarke coordinates} in honor of Edith Clarke.

\begin{figure}
    \centering
    \includegraphics[width=0.65\columnwidth]{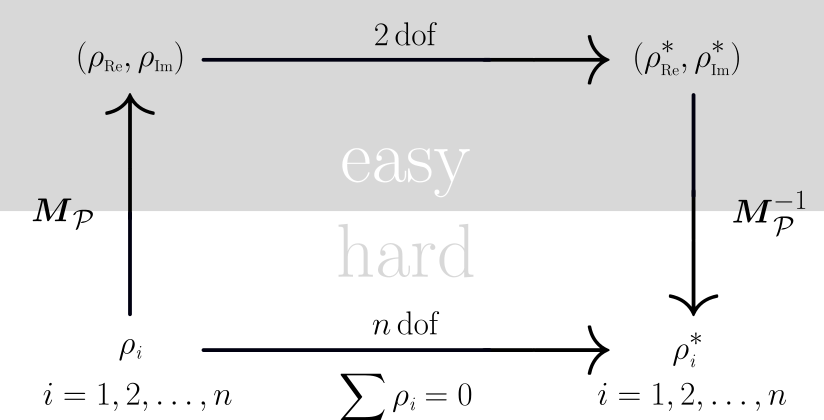}
    \caption{
    Commutative diagram-like overview.
    To circumvent the explicit consideration of the displacement constraint, \textit{i.e.}, $\sum \rho_i = 0$, approaches and methods should be considered on the \SI{2}{dof} manifold embedded in the $n\,$\SI{}{dof} joint space.
    For the transform, linear maps $\boldsymbol{M}_\mathcal{P}$ and $\boldsymbol{M}_\mathcal{P}^{-1}$ are used and any output denoted by $\left(\rho_\text{Re}^*, \rho_\text{Im}^*\right)$ can subsequently mapped back leading to $\rho^*$.
    }
    \label{fig:commutative_diagram}
\end{figure}

As shown in \eqref{eq:forward_mapping}, the Clarke transform is useful for converting $n$ displacements to two Clarke coordinates.
In this case, the Clarke transform simplifies to a simple matrix multiplication using $\boldsymbol{M}_\mathcal{P}$.
The Clarke coordinates can then be used as input and output in an approach.
Note that the dimensionality reduces from $n$ to two.
By using $\boldsymbol{M}_\mathcal{P}^{-1}$ as in \eqref{eq:inverse_mapping}, Clarke coordinates are transformed by \eqref{eq:inverse_mapping} to the $n$ displacements.
Inspired by depictions of the well-known Laplace transform and Fourier transform, a high-level illustration of the workflow is shown in Figure~\ref{fig:commutative_diagram}.

\subsection{Properties and Relations}

Here we state properties and relations, which might be useful when working with the $\MP$ and $\MPinv$.
Notes on their geometric interpretation and application are provided if appropriate.

From linear algebra, we can deduce a linear property, \textit{i.e.}, 
\begin{align}
    \MP
    \sum_{i=1}^{n}
    \rhovec_i
    = &
    \sum_{i=1}^{n}
    \MP
    \rhovec_i
    = 
    \sum_{i=1}^{n}
    \rhoclarke_i
    \quad\text{and}
    \\
    \MPinv
    \sum_{i=1}^{n}
    \rhoclarke_i
    = &
    \sum_{i=1}^{n}
    \MPinv
    \rhoclarke_i
    = 
    \sum_{i=1}^{n}
    \rhovec_i
    ,
\end{align}
and that $\MPinv$ is the right-inverse of $\MP$, \textit{i.e.},
\begin{align}
    \MP\MPinv = \Imat_{2 \times 2}
    ,
    \label{eq:MP_right_inverse}
\end{align}
whereas $\MPinv\MP \not= \Imat_{n \times n}$.
In fact, using the trigonometric identities and the index notation, this expression evaluates to
\begin{align}
    \MPinv\MP = \left( \dfrac{2}{n}\cos\left(2\pi\dfrac{i-j}{n}\right) \right)_{i,j} \in \mathbb{R}^{n \times n}
    ,
    \label{eq:MP_Toeplitz}
\end{align}
which is a symmetric circulant matrix being a special Toeplitz matrix.
For an engineer-friendly introduction to this topic, we kindly refer to a review by \cite{Gray_FTCIT_2006}.

A Toeplitz matrix like \eqref{eq:MP_Toeplitz} has two notable properties.
First, 
\begin{align}
    \left(\MPinv\MP\right)^k = \MPinv\MP
    \quad\text{for }k > 0
    \label{eq:MP_Toeplitz_idempotent}
\end{align}
shows that \eqref{eq:MP_Toeplitz} is an idempotent matrix.
\begin{proof}
    It is sufficient to show that $\left(\MPinv\MP\right)^2 = \MPinv\MP$.
    This simple case can be shown by squaring \eqref{eq:MP_Toeplitz} and applying trigonometric identities to simplify the expression.
\end{proof}

\noindent
Second, \eqref{eq:MP_Toeplitz} is a singular matrix and, therefore,
\begin{align}
    \det{\MPinv\MP} = 0
    \label{eq:MP_Toeplitz_det}
\end{align}
\begin{proof}
    This is evident by adapting the inverse of a nonsymmetric trigonometric Toeplitz matrix presented by \cite{Dow_ANZIAM_2002} to \eqref{eq:MP_Toeplitz_idempotent}.
    The inverse is singular if the nonsymmetric trigonometric Toeplitz matrix \citep{Dow_ANZIAM_2002} is symmetric, indicated by a zero division of the common factor. 
    The common factor is linear to the determinant of the matrix.
    Therefore, its determinant is zero.
    Moreover, it is known that the only non-singular idempotent matrix is the identity matrix, which is not the case because $\MPinv$ is the right-inverse, see \eqref{eq:MP_right_inverse}.
\end{proof}

Due to the similar entries in \eqref{eq:MP} and \eqref{eq:MP_inverse}, the relation
\begin{align}
    \MP\transpose = \dfrac{2}{n}\MPinv
    \label{eq:MP_transpose_relation}
\end{align}
holds.
This relation can be used to simplify an expression to $\Imat_{2 \times 2}$ by identifying \eqref{eq:MP_right_inverse} or to reformulate \eqref{eq:MP_Toeplitz_det} and \eqref{eq:MP_Toeplitz_det}.
\begin{proof}
    Given \eqref{eq:MP} and \eqref{eq:MP_inverse}, the relation can be written down by inspection.
\end{proof}

\begin{figure*}
    \centering
    \includegraphics[width=\textwidth]{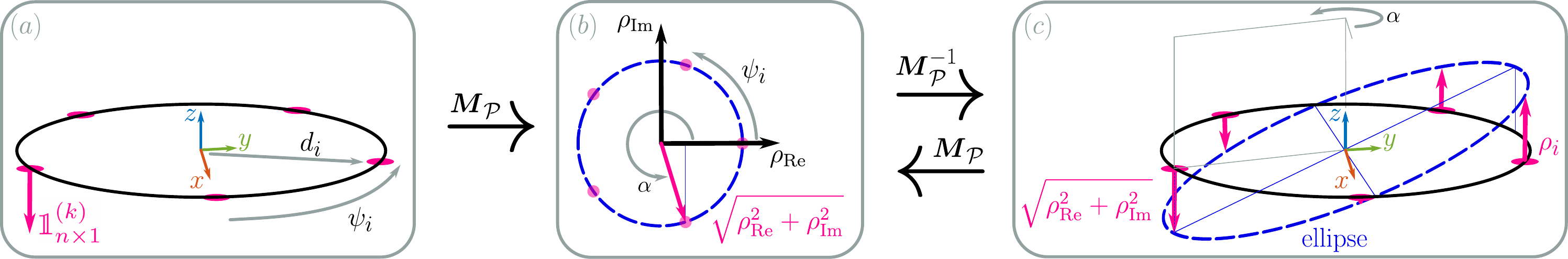}
    \caption{
        Visual aid for property \eqref{eq:properties_hotvec} and \eqref{eq:properties_circle_manifold}.
        Regarding the property \eqref{eq:properties_hotvec} illustrated in the sequence (a) to (c), a one-hot vector $\hotvec_{n \times 1}^{(k)}$ amplifies after mapping back from the manifold to the displacement.
        Its magnitude remains the same, whereas all other displacements are either equal or smaller than a previous one-hot vector.
        Regarding the property \eqref{eq:properties_circle_manifold} illustrated in (b) and (c), the $\rhoclarke$ in \eqref{eq:properties_circle_manifold} parameterized by $\alpha$ describes a circle.
        Tracing the tip of all possible displacement parameterized by $\psi \in \left[0, 2\pi\right)$ creates an ellipse.
        Its semi-major and semi-minor axes are $\sqrt{d^2 + \rhoreal^2 + \rhoim^2}$ and $d$ length, respectively.
        The maximum displacement achievable coincides with the angle $\alpha$, hinting at its geometric interpretation.
        }
    \label{fig:unit_circle_on_manifold}
\end{figure*}

Now, different vectors are evaluated when transformed by $\MP$ or $\MPinv$.
First, we can select a mode by  
\begin{align}
    \MP\hotvec_{n \times 1}^{(k)} 
    =
    \begin{bmatrix}
        \cos\left(\psi_k\right) \\
        \sin\left(\psi_k\right)
    \end{bmatrix}
    = 
    \begin{bmatrix}
        \cos\left(2\pi\left(k-1\right) / n\right) \\
        \sin\left(2\pi\left(k-1\right) / n\right)
    \end{bmatrix}
    ,
    \label{eq:properties_hotvec}
\end{align}
where $\hotvec_{n \times 1}^{(k)}$ is a one-hot vector defined by the $k\textsuperscript{th}$ element to be one, whereas all other elements are zero.
Figure~\ref{fig:unit_circle_on_manifold} provides a visual aid to the effect of the property \eqref{eq:properties_hotvec}.
Second, if $\rhovec \notin \mathbf{Q}$ has a bias term, then this bias term vanishing, \textit{i.e.},
\begin{align}
    \MP\onevec_{n \times 1}
    =
    \mathbf{0}_{2 \times 1},
    \label{eq:properties_vanishing_bias}
\end{align}
where $\onevec_{n \times 1}$ has ones everywhere.
\begin{proof}
    By applying \eqref{eq:properties_hotvec} for all entries, the left-hand side of \eqref{eq:properties_vanishing_bias} results in \eqref{eq:sum_cosine} and \eqref{eq:sum_sine}, respectively.
    Hence, the right-hand side of \eqref{eq:properties_vanishing_bias} is a zero vector.
\end{proof}
\noindent
Third, a unit circle can be described on the manifold by parametrizing $\rhoclarke$ using $\alpha \in \left[0, 2\pi\right)$.
Its Clarke transform is
\begin{align}
    \MPinv
    \begin{bmatrix}
        \cos\left(\alpha\right)\\
        \sin\left(\alpha\right)
    \end{bmatrix}
    = 
    \begin{bmatrix}
        \cos\left(\psi_1 - \alpha\right)\\
        \cos\left(\psi_2 - \alpha\right)\\
        \vdots \\
        \cos\left(\psi_n - \alpha\right)
    \end{bmatrix}
    .
    \label{eq:properties_circle_manifold}
\end{align}
Note that the relations \eqref{eq:properties_hotvec}, \eqref{eq:properties_vanishing_bias}, and \eqref{eq:properties_circle_manifold} are not invertible due to \eqref{eq:MP_Toeplitz} and \eqref{eq:MP_Toeplitz_det}.

The property \eqref{eq:properties_circle_manifold} provides geometric insights.
Assume infinite many displacement-actuated joints such that $\psi_i$ becomes $\psi \in \left[0, 2\pi\right)$.
One of the differences $\psi - \alpha$ evaluates to zero, and another evaluates to $\pi$. 
Therefore, the right-hand side of \eqref{eq:properties_circle_manifold} has one value that evaluates to one, and another that evaluates to minus one being the maximum and minimum, respectively.
Figure~\ref{fig:unit_circle_on_manifold} provides visual aid.

Furthermore, \eqref{eq:properties_circle_manifold} underlines the intuition we gained from $\MP\left(n = 4\right)$. 
For this, $\alpha = 0$ aligns with first actuation location, \textit{i.e.}, $\psi_1 = 0$, and $\rho_1 = 1$ is maximum of $\rhovec$.
Therefore, $\rhoreal$ aligns with the $x$-axis of the base frame.
A similar argument can be made for $\rhoim$ that aligns with the $y$-axis of the base frame.

For simplification purposes, the following relations are evident, yet useful.
Each of the Clarke coordinates can be extracted from \eqref{eq:forward_mapping} using a selection matrix, \textit{i.e.},
\begin{align}
    \rhoreal = \begin{bmatrix} 1 & 0 \end{bmatrix}
    \MP\rhovec
    \quad\text{and}\quad
    \rhoim = \begin{bmatrix} 0 & 1 \end{bmatrix}
    \MP\rhovec
    \label{eq:rhoclarke_selection}
\end{align}
Furthermore, the sum of squares is the dot product given by
\begin{align}
    \rhoreal^2 + \rhoim^2
    =
    \rhoclarke\transpose\rhoclarke
    = 
    \rhovec\transpose\MP\transpose\MP\rhovec
    .
    \label{eq:rhoclarke_magnetude_MP}
\end{align}
By using \eqref{eq:inverse_mapping}, the sum of squares results in 
\begin{align}
    \rhoclarke\transpose\rhoclarke
    = 
    \dfrac{2}{n}\rhovec\transpose\rhovec
    .
    \label{eq:rhoclarke_magnetude_2n}
\end{align}
\section{Physical Meaning of the Clarke Coordinates}
\label{sec:interpretation}

Building on the intuition gained from $\MP\left(n = 4\right)$ and the property \eqref{eq:properties_circle_manifold}, this section investigates the physical meaning of the Clarke coordinates.
By doing this, arc parameters used for models with constant curvature assumptions are considered to derive the connection.

\subsection{Relation to arc parameters}

One approach to explore the physical meaning of $\rho_\mathrm{Re}$ and $\rho_\mathrm{Im}$ is to relate the displacement formulation \eqref{eq:rho_rectangular_form} to known representations in the literature.
Instead of displacement $\rho_i$, in the literature, \textit{e.g.}, \citep{WebsterJones_IJRR_2010, RaoBurgner-Kahrs_et_al_Frontiers_2021}, the absolute length $l_i$ is frequently used.
Using our notation, one can state 
\begin{align}
    l = l_i + \phi d \cos\left(\psi_i - \theta\right)
    ,
    \label{eq:tendon_length_literature}
\end{align}
where $l$ is the length in the straight configuration, which is equal to the segment length.
Note that $\theta$ and $\phi$ are one set of arc parameters, see Appendix~\ref{appendix:arc_space_representation}.
Recap, $\psi_i$ and $d_i = d$ describe the $i\textsuperscript{th}$ location \eqref{eq:psi}.
To reformulate \eqref{eq:tendon_length_literature} as displacement, the well-known fact that $\rho_i = l - l_i$ and $\phi = l\kappa$ are used leading to
\begin{align}
    \rho_i = dl\kappa\cos\left(\psi_i - \theta\right)
    .
    \label{eq:rho_literature}
\end{align}
A key approach is again the use of the harmonic addition \eqref{eq:harmonic_addition}.
With that, we can identify the structure of displacement in the rectangular form \eqref{eq:rho_rectangular_form} and write
\begin{align}
    \rho_i = dl\kappa\cos(\psi_i) \cos(\theta) + dl\kappa\sin(\psi_i) \sin(\theta).
    \nonumber
\end{align}
By comparing the above equation to \eqref{eq:rho_rectangular_form} or identifying \eqref{eq:harmonic_addition_A_B}, the physical meaning is revealed, assuming constant curvature.
As a consequence, Clarke coordinates can be described by
\begin{align}
    \rho_\mathrm{Re} = d l \kappa \cos\left(\theta\right)
    % \label{eq:rho_re_fdep}
    % \\
    \quad\text{and}\quad
    \rho_\mathrm{Im} = d l \kappa \sin\left(\theta\right)
    .
    % \label{eq:rho_im_fdep}
    \label{eq:rho_2dof_fdep_from_car}
\end{align}
Thus $\rho_\mathrm{Re}$ and $\rho_\mathrm{Im}$ in \eqref{eq:rho_2dof_fdep_from_car} are related to another representation of the arc space, see Appendix~\ref{appendix:arc_space_representation}.
This representation of the arc space is given by arc parameters
\begin{align}
	 \kappa_x  = \kappa\cos(\theta) \quad\text{and}\quad \kappa_y  = \kappa\sin(\theta)
  \label{eq:kappa_xy_from_car}
\end{align}
combining the arc parameters $\kappa$ and $\theta$ in a non-linear way.

\subsection{Virtual displacement}
\label{sec:virtual_displacement}

\begin{figure}
    \centering
    \includegraphics[width=0.55\columnwidth]{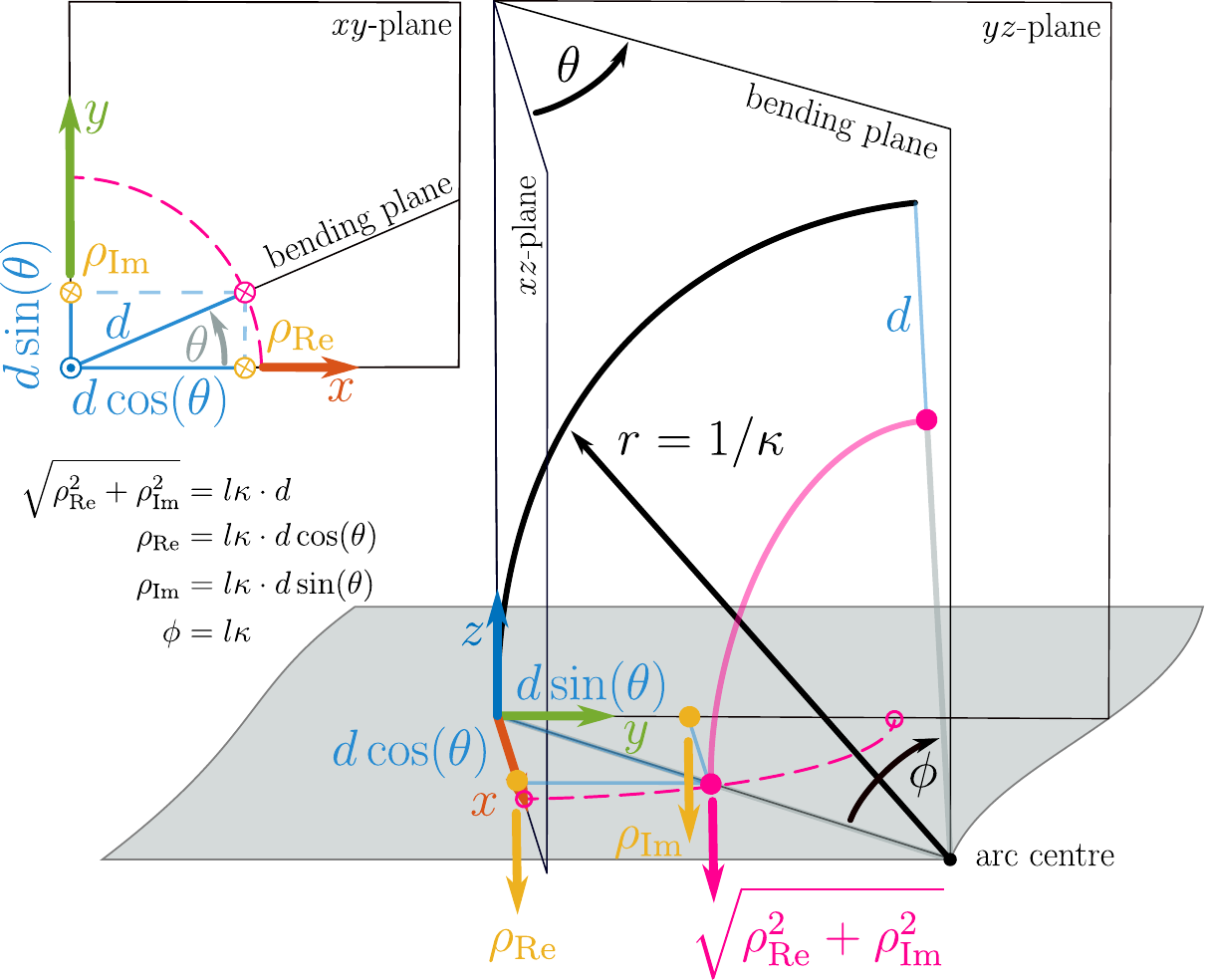}
    \caption{
        Physical interpretation of the Clarke coordinates.
        The magenta line lies within the bending plane.
        The length difference to the arc length is the virtual displacement.
        The virtual displacement is equal to $\sqrt{\rho_\text{Re}^2 + \rho_\text{Im}^2} = dl\kappa = d\phi$.
        The yellow arrows lie within the respective projected plane corresponding to the $xz$-plane and $yz$-plane of the base.
        Similar to the curvature components $\kappa_x$ and $\kappa_y$, the virtual displacement can be projected onto the respective plane, resulting in the respective projected virtual displacement.
    }
    \label{fig:rho_physical_interpretation}
\end{figure}

To further gain insights into the physical meaning, we utilize the relationship to harmonic addition \eqref{eq:harmonic_addition}.
By exploiting \eqref{eq:harmonic_addition_D_theta} or directly comparing \eqref{eq:rho_literature} and polar form \eqref{eq:rho_polar_form}, the equations
\begin{align}
    \sqrt{\rho_\text{Re}^2 + \rho_\text{Im}^2} = dl\kappa \quad\text{and}\quad
    \arctantwo\left({\rho_\text{Im},\ \rho_\text{Re}}\right) = \theta
    \label{eq:rho_polar_form_ccr}
\end{align}
can be found, where $d\phi$ can be used instead of $dl\kappa$.
By finding \eqref{eq:rho_2dof_fdep_from_car} and \eqref{eq:kappa_xy_from_car}, the relation to the arc space is established.
Due to the fact that arc parameters relate to the constant curvature assumption \citep{WebsterJones_IJRR_2010}, the Clarke coordinates can be interpreted geometrically.
Furthermore, $\sqrt{\rho_\text{Re}^2 + \rho_\text{Im}^2}$ stated in \eqref{eq:rho_polar_form_ccr} is equivalent to virtual tendon length introduced by \cite{FirdausVadali_AIR_2023}.
The virtual tendon is conceptional a displacement, which lies in the bending plane rotated by $\theta$ and is equal to the maximum displacement $dl\kappa = d\phi$.
Therefore, $\rho_\mathrm{Re}$ and $\rho_\mathrm{Im}$ are the projection of $dl\kappa$ onto the $xz$-plane and $yz$-plane, respectively, \textit{cf.} \eqref{eq:kappa_xy_from_car}.
Figure~\ref{fig:rho_physical_interpretation} illustrated this.
The scaling of radius $d$ is relevant, which leads to $d_x = d\cos\left(\theta\right)$ and $d_y = d\sin\left(\theta\right)$.

The \textit{virtual displacement} is the maximum displacement coincided with the bending plane and results in 
\begin{align}
    \sqrt{\rho_\text{Re}^2 + \rho_\text{Im}^2} = l\kappa \cdot d = \phi \cdot d,
    \label{eq:virtual_displacement}
\end{align}
whereas the \textit{projected virtual displacements} given by
\begin{align}
    \rho_\mathrm{Re} &= \phi \cdot d_x = l\kappa \cdot d_x
    \quad\text{and}\quad
    \label{eq:virtual_displacement_x}
    \\
    \rho_\mathrm{Im} &= \phi \cdot d_y = l\kappa \cdot d_y
    \label{eq:virtual_displacement_y}
\end{align}
are projections of \eqref{eq:virtual_displacement}.
Therefore, $\rho_\mathrm{Re}$ and $\rho_\mathrm{Im}$ are primarily displacements and not arc parameters.
Note that \eqref{eq:virtual_displacement}, \eqref{eq:virtual_displacement_x}, and \eqref{eq:virtual_displacement_y} are direct consequences of \eqref{eq:rho_2dof_fdep_from_car}, \eqref{eq:rho_polar_form_ccr}, and $\phi = l\kappa$ with inspiration from \eqref{eq:kappa_xy_from_car}.

\section{Application to Kinematics}
\label{sec:kinematics}

In this section, we utilize the geometric insights from the previous section and derive the closed-form kinematics for constant curvature robots with one segment.
For constant curvature kinematics\citep{WebsterJones_IJRR_2010}, the robot-independent mapping $f_\text{ind}$ relates arc space to task space as summarized in Appendix~\ref{appendix:robot_independent_mapping}.
Given $f_\text{ind}$, solutions for the inverse robot-independent mapping $f_\text{ind}^{-1}$ are listed in Table~\ref{tab:solution_find_inv} for common arc space representations.
We derive generalised versions of the inverse and forward robot-dependent mappings using the Clarke transform in the following, \textit{i.e.}, mapping between the joint space and the arc space.
Afterwards, the mappings are combined to derive the forward and inverse kinematics.
This section ends with a brief discussion on singularities.

\begin{table*}
    \renewcommand*{\arraystretch}{1.4}
    \caption{
    Closed-form solutions for the inverse robot-independent mapping in a nutshell.
    }
    \label{tab:solution_find_inv}
    \centering
    \begin{tabular}{r r r r} 
        \toprule
        \multicolumn{1}{N}{arc parameter} & \multicolumn{1}{N}{$E\!\left(3\right)$} & \multicolumn{1}{N}{$SO\!\left(3\right)$} & \multicolumn{1}{N}{$SE\!\left(3\right)$} \\
		\cmidrule(r){1-1}
		\cmidrule(lr){2-2}
		\cmidrule(lr){3-3}
		\cmidrule(l){4-4}
        $\theta$ & $\arctantwo\left(p_y, p_x \right)$ & $\arctantwo\left(-r_{12}, r_{22}\right)$ & $\arctantwo\left(p_y, p_x \right)$ \\[.25em]
        $\kappa$ & $2\sqrt{p_x^2 + p_y^2} / \left( p_x^2 + p_y^2 + p_z^2 \right)$ & $\left(1 / l \right)\arctantwo\left(-r_{31}, r_{33}\right)$ & $-r_{31} / p_z$ \\[.25em]
        $\kappa_x$ & $ 2p_x / \left(p_x^2 + p_y^2 + p_z^2\right)$ & $\left(r_{22} / l\right)\arctantwo\left(-r_{31}, r_{33}\right)$ & $- r_{31}r_{22} / p_z$ \\[.25em]
        $\kappa_y$ & $2p_y / \left(p_x^2 + p_y^2 + p_z^2\right)$ & $\left(r_{11} / l\right) \arctantwo\left(-r_{31}, r_{33}\right)$ & $- r_{31}r_{11} / p_z$ \\[.25em]
        \bottomrule
    \end{tabular}
\end{table*}

\subsection{Robot-Dependent Mappings using Clarke Transform}
\label{sec:robot_dependent_mapping}

Using $\phi = l\kappa$, the formulation \eqref{eq:kappa_xy_from_car} is regarded as the inverse robot-dependent mapping $f_\text{dep}^{-1}$, which can be used for arbitrary $n$ displacement \citep{DalvandNahavandiHowe_TRO_2018, DalvandNahavandiHowe_Access_2022}.
It requires the evaluation of $n$ transcendental functions for $n$ displacements.
Furthermore, directly inverting $n$ non-linear functions to obtain the arc parameters is not straightforward and may not even be possible.
So far, the forward robot-dependent mapping $f_\text{dep}$ for an arbitrary number $n$ remains unknown.

\subsubsection{Inverse Robot-Dependent Mapping}
\label{sec:inverse_robot_dependent_mapping}

Formulating $f_\text{dep}^{-1}$ by utilizing \eqref{eq:inverse_mapping} and \eqref{eq:rho_2dof_fdep_from_car}, a more compact and computationally efficient approach can be formulated.
This leads to a $f_\text{dep}^{-1}$ expressed by 
\begin{align}
    \boldsymbol{\rho}
    = 
    d l \boldsymbol{M}_\mathcal{P}^{-1}
    \begin{bmatrix} 
        \kappa \cos\left(\theta\right)\\
        \kappa \sin\left(\theta\right)
    \end{bmatrix}
    = 
    d l \boldsymbol{M}_\mathcal{P}^{-1}
    \begin{bmatrix} 
        \kappa_x\\
        \kappa_y
    \end{bmatrix}
    ,
    \label{eq:rho_fdep_inverse}
\end{align}
where only two transcendental functions for $n$ displacements are evaluated, \textit{i.e.}, $\cos\left(\theta\right)$ and $\sin\left(\theta\right)$, when $\kappa$ and $\theta$ are used to represent the arc space.
Note that $\psi_i$ is a design parameter and, therefore, $\psi_i$ and $\boldsymbol{M}_\mathcal{P}^{-1}$ are constant.
When using $\kappa_x$ and $\kappa_y$, \eqref{eq:rho_fdep_inverse} is a linear mapping.
More importantly, $\boldsymbol{M}_\mathcal{P}^{-1}$ is invertible, which is $\boldsymbol{M}_\mathcal{P}$.
This opens the possibility to formulate a forward robot-dependent mapping $f_\text{dep}$.

\subsubsection{Forward Robot-Dependent Mapping}
\label{sec:forward_robot_dependent_mapping}

Using the fact that $\boldsymbol{M}_\mathcal{P}^{-1}$ is the right-inverse \eqref{eq:MP_right_inverse}, the inverse robot-dependent mapping \eqref{eq:rho_fdep_inverse} can be inverted leading to
\begin{align}
    \begin{bmatrix}
        \kappa_x \\ \kappa_y
    \end{bmatrix}
    = 
    \begin{bmatrix}
        \kappa\cos\left(\theta\right) \\ \kappa\sin\left(\theta\right)
    \end{bmatrix}
    = 
    \dfrac{1}{dl}
    \boldsymbol{M}_\mathcal{P}
    \boldsymbol{\rho}
    .
    \label{eq:rho_fdep}
\end{align}
Hence, by utilizing the Clarke transform, for the first time, a closed-form forward and inverse robot-dependent mapping for a single-segment constant curvature robot with $n$ displacement actuations can be formulated.
Both formulations for $f_\text{dep}$ and $f_\text{dep}^{-1}$ highlight that the curvature-curvature representation can be seen as the desired parameterization of the arc space thanks to their linear relation to $\rhovec$.

For the curvature-angle representation, we can sum the squares of \eqref{eq:rho_2dof_fdep_from_car} in two different ways.
First, using \eqref{eq:virtual_displacement} and solving for $\kappa^2$ results in $\kappa^2 = \left( \rho_\text{Re}^2 + \rho_\text{Im}^2 \right) / \left( dl \right)^2$.
Then, \eqref{eq:rhoclarke_magnetude_MP} and \eqref{eq:rhoclarke_magnetude_2n} are used.
As a result, the curvature is
\begin{align}
    \kappa = \dfrac{1}{dl}\sqrt{\rhovec\transpose\MP\transpose\MP\rhovec} = \dfrac{\sqrt{2n}}{dln}\sqrt{\rhovec\transpose\rhovec}.
    \label{eq:kappa_fdep_car}
\end{align}
By considering \eqref{eq:rhoclarke_selection} and \eqref{eq:rho_polar_form_ccr}, we can state
\begin{align}
    \theta = \arctantwo\left(\left[0, 1\right]\boldsymbol{M}_\mathcal{P}\boldsymbol{\rho},\ \left[1, 0\right]\boldsymbol{M}_\mathcal{P}\boldsymbol{\rho}\right)
    \label{eq:theta_fdep_car}
\end{align}
completing the derivation of the $f_\text{dep}$.

\subsection{Closed-form Kinematics}
\label{sec:kinematics_closed_form}

\begin{figure}
    \centering
    \includegraphics[width=0.55\columnwidth]{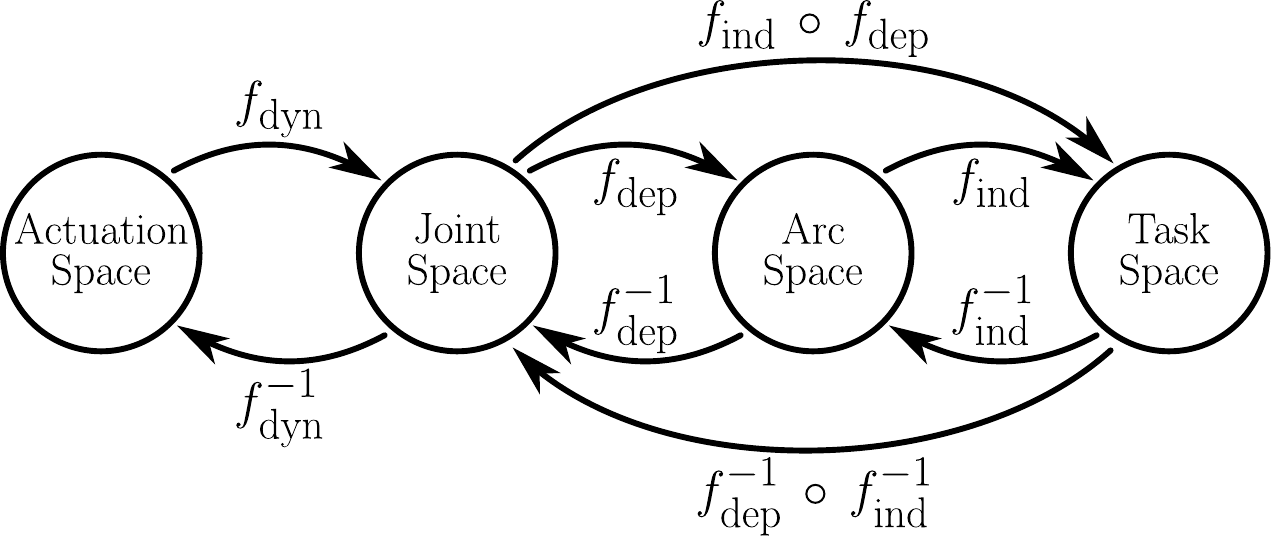}
    \caption{
    Spaces and their mappings.
    To map between the actuator, \textit{e.g.}, motor angles to joint values, \textit{e.g.}, tendon displacements, $f_\text{dyn}$ and $f_\text{dyn}^{-1}$ are used.
    Note that it is crucial to treat actuation space and joint space as different spaces as $f_\text{dyn}$ is not necessarily a linear map. 
    The robot-dependent mapping is denoted by $f_\text{dep}$, whereas robot-independent mapping is denoted by $f_\text{ind}$.
    Given both mappings, the forward and inverse kinematics can be found using $f_\text{ind}\circ f_\text{dep}$ and $f_\text{dep}^{-1}\circ f_\text{ind}^{-1}$, respectively.
    }
    \label{fig:spaces_spaces}
\end{figure}

Utilizing the \SI{2}{dof} manifold, we can directly state the forward kinematics $f_\text{ind}\circ f_\text{dep}$ and the inverse kinematics $f_\text{dep}^{-1}\circ f_\text{ind}^{-1}$ without evaluating the arc parameters. 
Figure~\ref{fig:spaces_spaces} illustrates this.

\subsubsection{Forward Kinematics}

Computing the forward kinematics $f_\text{ind}\circ f_\text{dep}$ can be reduced to the evaluation of the trigonometric functions $\cos\left(\theta\right)$, $\sin\left(\theta\right)$, $\cos\left(\kappa l\right)$, and $\sin\left(\kappa l\right)$, as well as the inverse of the curvature $\kappa$, see the classic formulation stated in Appendix~\ref{appendix:robot_independent_mapping}.
The displacement $\rhovec$ is the input of forward kinematics $f_\text{ind} \circ f_\text{dep}$.

Using \eqref{eq:kappa_fdep_car}, the inverse of the curvature is described by
\begin{align}
    1 / \kappa = \dfrac{dl\sqrt{2n}}{2\sqrt{\rhovec\transpose\rhovec}}.
    \label{eq:kappa_inv_fdep}
\end{align}
The angle $\phi = \kappa l$ uses \eqref{eq:kappa_fdep_car}, 
where $d > 0$, and, therefore, $\kappa l$ is well-defined meaning it cannot be singular.
Now, we can formulate
\begin{align}
    \cos\left(\kappa l\right) &= \cos\left(\dfrac{\sqrt{2n}}{dn}\sqrt{\rhovec\transpose\rhovec}\right)\quad\text{and}\\
    \sin\left(\kappa l\right) &= \sin\left(\dfrac{\sqrt{2n}}{dn}\sqrt{\rhovec\transpose\rhovec}\right).
\end{align}
Using components of \eqref{eq:rho_fdep} and inserting \eqref{eq:kappa_fdep_car}, we can derive
\begin{align}
    \cos\left(\theta\right) &= \left(\dfrac{2}{n}\rhovec\transpose\rhovec\right)^{-1/2} \left[1, 0\right]\boldsymbol{M}_\mathcal{P}\boldsymbol{\rho}
    \quad\text{and}
    \label{eq:cos_theta_direct_fk}
    \\
    \sin\left(\theta\right) &= \left(\dfrac{2}{n}\rhovec\transpose\rhovec\right)^{-1/2}\left[0, 1\right]\boldsymbol{M}_\mathcal{P}\boldsymbol{\rho}
    .
    \label{eq:sin_theta_direct_fk}
\end{align}
Substituting the equations above into classic formulation, \textit{i.e.}, \eqref{eq:p_indep} and \eqref{eq:R_indep} in Appendix~\ref{appendix:robot_independent_mapping}, completes the forward kinematics $f_\text{ind}\circ f_\text{dep} : \left(\boldsymbol{\rho}\right)\rightarrow \left(\boldsymbol{p}, \boldsymbol{R}\right)$.

\subsubsection{Inverse Kinematics}

The relation between the Clarke coordinates and the curvature-curvature presentation given in \eqref{eq:rho_2dof_fdep_from_car} is instrumental in developing the inverse kinematics $f_\text{dep}^{-1} \circ f_\text{ind}^{-1}$.
The solutions are listed in Table~\ref{tab:solution_ik} for different sources of information expressed in task space, \textit{i.e.}, $E\!\left(3\right)$, $SO\!\left(3\right)$, or $SE\!\left(3\right)$.

\begin{table*}
    \renewcommand*{\arraystretch}{1.4}
    \caption{
    Closed-form solutions for the inverse kinematics in a nutshell.
    }
    \label{tab:solution_ik}
    \centering
    \begin{tabular}{r r r r} 
        \toprule
        \multicolumn{1}{N}{$\mathbf{Q}$} & \multicolumn{1}{N}{$E\!\left(3\right)$} & \multicolumn{1}{N}{$SO\!\left(3\right)$} & \multicolumn{1}{N}{$SE\!\left(3\right)$} \\
		\cmidrule(r){1-1}
		\cmidrule(lr){2-2}
		\cmidrule(lr){3-3}
		\cmidrule(l){4-4}
        \\[-0.875em]
        $\rhovec$ & 
        $\dfrac{2dl}{p_x^2 + p_y^2 + p_z^2}\MPinv\begin{bmatrix} p_x & p_y \end{bmatrix}\transpose$ & 
        $d \arctantwo\left(-r_{31}, r_{33}\right)\MPinv\begin{bmatrix} r_{22} & r_{12} \end{bmatrix}\transpose$ & 
        $\dfrac{-dlr_{31}}{p_z}\MPinv\begin{bmatrix} r_{22} & r_{12} \end{bmatrix}\transpose$ \\[1em]
        \bottomrule
    \end{tabular}
\end{table*}

\paragraph{Tip Position}
Expressing \eqref{eq:rho_2dof_fdep_from_car} using \eqref{eq:kappa_xy_from_car} as well as $\kappa_x$ and $\kappa_y$ stated for $E\!\left(3\right)$ in Table~\ref{tab:solution_find_inv} leads to two equations.
Both equations differ by $p_x$ and $p_y$, respectively.
The combination
\begin{align}
  \rhovec = 
  \dfrac{2dl}{p_x^2 + p_y^2 + p_z^2}\MPinv
  \begin{bmatrix}
      p_x \\
      p_y
  \end{bmatrix}
\end{align}
is the inverse kinematics for a given position, where \eqref{eq:inverse_mapping} is used.

\paragraph{Tip Orientation}
Starting with \eqref{eq:rho_2dof_fdep_from_car} with \eqref{eq:kappa_xy_from_car}, we can use $\kappa_x$ and $\kappa_y$ stated for $SO\!\left(3\right)$ in Table~\ref{tab:solution_find_inv} and \eqref{eq:inverse_mapping} to find
\begin{align}
  \rhovec = 
  d \arctantwo\left(-r_{31}, r_{33}\right)\MPinv
  \begin{bmatrix}
      r_{22} \\ r_{12}
  \end{bmatrix}
  ,
\end{align}
where the $l$ cancels out.
Note that the tip orientation alone cannot be used to find the arc length $l$, see Appendix~\ref{appendix:robot_independent_mapping}.

\paragraph{Tip Pose}
By inserting $\kappa_x$ and $\kappa_y$ stated for $SE\!\left(3\right)$ in Table~\ref{tab:solution_find_inv} into \eqref{eq:rho_2dof_fdep_from_car} with \eqref{eq:kappa_xy_from_car}, and, afterwards, utilizing \eqref{eq:inverse_mapping},
\begin{align}
  \rhovec = 
  \dfrac{-dlr_{31}}{p_z}\MPinv
  \begin{bmatrix}
      r_{22} \\ r_{12}
  \end{bmatrix}
\end{align}
can be derived.
Note that both lines differ by $r_{22}$ and $r_{21}$, respectively.

\subsection{Singularities in the Representation}
\label{sec:singularities}

It is well-known that the mathematical formulation of $f_\text{ind}$ using the curvature-angle representation can become singular.
In the classic formulation \citep{WebsterJones_IJRR_2010} of $f_\text{ind}$ stated in Appendix~\ref{appendix:robot_independent_mapping}, the expressions
\begin{align}
    \left(1 - \cos(\kappa l)\right) / \kappa
    \quad\text{and}\quad
    \sin(\kappa l) / \kappa
    \label{eq:sinc_1cosxx}
\end{align}
are troublesome and will lead to an indeterminate form as $\kappa$ approaches zero.
Although not often mentioned, ad-hoc solutions do exist due to their limits being well-defined, \textit{i.e.},
\begin{align}
    \lim_{\kappa\rightarrow \pm 0}\dfrac{1 - \cos(\kappa l)}{\kappa} = 0
    \quad\text{and}\quad
    \lim_{\kappa\rightarrow \pm 0}\dfrac{\sin(\kappa l)}{\kappa} = l.
    \label{eq:limits_sinc_1cosxx}
\end{align}
One might simply replace \eqref{eq:sinc_1cosxx} at $\kappa = 0$ with \eqref{eq:limits_sinc_1cosxx}, which is used to derive \eqref{eq:p_indep_straight}.

Using a threshold $\epsilon > 0$ instead, one might switch to their truncated expansions before \eqref{eq:sinc_1cosxx} becomes undefined.
For this, the Taylor series of \eqref{eq:sinc_1cosxx} can be used, which are
\begin{align}
    \dfrac{\sin(\kappa l)}{\kappa} &= l - \dfrac{l^3\kappa^2}{3!} + \dfrac{l^5\kappa^4}{5!} - \dfrac{l^7\kappa^6}{7!} + \cdots
    \quad\text{and}
    \nonumber
    \\
    \dfrac{1 - \cos(\kappa l)}{\kappa} &= \dfrac{l^2\kappa}{2!} - \dfrac{l^4\kappa^3}{4!} + \dfrac{l^6\kappa^5}{6!} + \cdots,
    \nonumber
\end{align}
respectively.
This is a known approach, \textit{e.g.}, \citep{RolfSteil_IROS_2012}, and relies on branching for the switching between \eqref{eq:limits_sinc_1cosxx} and the truncated Taylor series.
The truncation depends on the validity of the neighborhood of that threshold $\epsilon$.

In contrast, the forward kinematics $f_\text{ind}\circ f_\text{dep}$ can circumvent the coordinate singularity of \eqref{eq:sinc_1cosxx}.
However, the normalization of actuation displacement $\boldsymbol{\rho}$ is used and can cause numerical issues if it approaches zero.
Therefore, to avoid division by zero in \eqref{eq:kappa_inv_fdep}, \eqref{eq:cos_theta_direct_fk}, and \eqref{eq:sin_theta_direct_fk}, we can reformulate them by adding a sufficient small $\epsilon > 0$, \textit{i.e.}, 
\begin{align}
    \sqrt{\rhovec\transpose\rhovec}\ \longleftarrow \sqrt{\rhovec\transpose\rhovec} + \epsilon.
    \label{eq:rho_norm_epsilon}
\end{align}
This approach avoids the use of branching, which, if at all, is normally used for curvature-curvature representation.
Assuming $\epsilon^2 \approx 0$, the introduced error is linear \textit{w.r.t.} $\epsilon$.
Furthermore, \eqref{eq:rho_norm_epsilon} is easy to implement.
As an alternative to avoid singularities, the \textsc{Fast Inverse Square Root} algorithm or other algorithms, 
performing a division and a root using only multiplication, can be used for \eqref{eq:cos_theta_direct_fk}, \eqref{eq:sin_theta_direct_fk}, and \eqref{eq:kappa_inv_fdep}.
\section{Application to Joint Space Sampling}
\label{sec:sampling}

In this section, we demonstrate the benefits of the Clarke transformations over existing representations in joint space sampling.
We evaluate two common rejection sampling methods (referred to as (a) and (b) below) against those based on $\boldsymbol{M}_\mathcal{P}^{-1}$ (referred to as (c) and (e) below), focusing on three joints uniformly distributed with radius $d = \SI{1}{mm}$. 
The Monte Carlo method is utilized to analyze the distribution of $\rho_i$ in the range $\left[ \rho_\text{min}, \rho_\text{max}\right] = \left[-d\pi, d\pi\right]$ and to quantify computational load.

\begin{table*}[t]
    \caption{
        Computational load regarding wall-clock time and number of iterations for different sampling methods.
        Each method is run five times, generating \num{1000} samples.
        The mean value and standard deviation are reported for the wall-clock time and number of iterations.
        The stated factor is the normalized wall-clock time \textit{w.r.t.} the direct sampling method.
        The number of iterations is the number of samples in a while loop.
    }
    \label{tab:sampling_comparision}
    \setlength{\tabcolsep}{4pt}
    \renewcommand{\arraystretch}{1.4}
    \centering
    \begin{tabular}{r r r r r r r r} 
        \toprule
        method  & vectorized & $\boldsymbol{M}_\mathcal{P}^{-1}$ & time in \SI{}{ms} & factor & iteration & resample & success rate in \%\\
		\cmidrule(r){1-1}
		\cmidrule(lr){2-2}
		\cmidrule(lr){3-3}
		\cmidrule(lr){4-4}
		\cmidrule(lr){5-5}
		\cmidrule(lr){6-6}
		\cmidrule(lr){7-7}
		\cmidrule(l){8-8}
        (a) & \checkNo & \checkNo & \num{258.9 \pm 17.2}  & \num{172.6} & \num{827440.2 \pm 60183.2}  & \num{826440} & \num{0.0012}\\
        (b) & \checkNo & \checkNo & \num{1.9 \pm 1.6}  & \num{1.3} & \num{1344.0 \pm 11.8}  & \num{1244} & \num{0.744}\\
        (c) & \checkNo & \checkYes & \num{1.5 \pm 0.8}  & \num{1} & \num{1000} & \num{0} & \num{100} \\
        (d) & \checkNo & \checkYes & \num{1.4 \pm 0.7}  & \num{0.9} & \num{1000} & \num{0} & \num{100} \\
        (e) & \checkNo & \checkYes & \num{1.4 \pm 0.7}  & \num{0.9} & \num{1000} & \num{0} & \num{100} \\
		\cmidrule(r){1-1}
		\cmidrule(lr){2-2}
		\cmidrule(lr){3-3}
		\cmidrule(lr){4-4}
		\cmidrule(lr){5-5}
		\cmidrule(lr){6-6}
		\cmidrule(lr){7-7}
		\cmidrule(l){8-8}
        (c) & \checkYes & \checkYes & \num{0.7 \pm 0.6}  & \num{0.5} & \num{1} & \num{0} & \num{100} \\
        (d) & \checkYes & \checkYes & \num{0.6 \pm 0.6}  & \num{0.4} & \num{1} & \num{0} & \num{100} \\
        (e) & \checkYes & \checkYes & \num{0.6 \pm 0.6}  & \num{0.4} & \num{1} & \num{0} & \num{100} \\
        \bottomrule
    \end{tabular}
\end{table*}

\subsection{Rejection Sampling via Branching}
\label{sec:rejection_sampling_via_branching}

One of two rejection sampling methods is used to sample the $k\textsuperscript{th}$ sample.
To avoid directly comparing float numbers, we round random values to two decimal places, setting $\epsilon = \SI{0.01}{mm}$.

(a) For the first rejection sampling method, all joint values for the $k\textsuperscript{th}$ sample are sampled independently, \textit{i.e.},
\begin{align}
    \rho_i^{\left(k\right)} &= \rho_\text{min} + \left(\rho_\text{max} - \rho_\text{min}\right)\mathcal{U}\left[0; 1\right]
    \label{eq:sampling_vanilla}
\end{align}
for $i \in \{1, 2, 3$\}, where $\mathcal{U}\left[0; 1\right]$ is the uniform distribution.
The sample is then tested against the displacement constraint \eqref{eq:sum_rho} and resampled if it fails to meet the criterion.

(b) The second rejection sampling method utilizes \eqref{eq:sum_rho} to compute $\rho_1$, \textit{i.e.}, $\rho_1 = - \rho_2 - \rho_3$.
For $\rho_2$ and $\rho_3$, \eqref{eq:sampling_vanilla} is used.
This method can be expressed as
\begin{flalign}
    \begin{bmatrix}
        \rho_1^{\left(k\right)} \\[0.25em]
        \rho_2^{\left(k\right)} \\[0.25em]
        \rho_3^{\left(k\right)}
    \end{bmatrix}
    \!=\!
    \begin{bmatrix} 
        -1 & -1 \\
         1 &  0 \\
         0 & 1 
    \end{bmatrix}
    \!
    \begin{bmatrix} 
        \rho_\text{min} + \left(\rho_\text{max} - \rho_\text{min}\right)\mathcal{U}\left[0; 1\right]\\
        \rho_\text{min} + \left(\rho_\text{max} - \rho_\text{min}\right)\mathcal{U}\left[0; 1\right]
    \end{bmatrix}.
    \label{eq:sampling_rho_1}
\end{flalign}
A sample generated by \eqref{eq:sampling_rho_1} might be resampled, since $\rho_1$ does not necessary comply with given boundaries, \textit{i.e.}, $\rho_\text{min}$ and $\rho_\text{max}$.
Therefore, vectorization of \eqref{eq:sampling_rho_1} has no computational advantage.

Note that permutation of the rows in \eqref{eq:sampling_rho_1} yields the other two possible sampling methods utilizing \eqref{eq:sum_rho}. 
However, distributions and computational load are similar and, therefore, not further considered.

\subsection{Direct Sampling}
\label{sec:direct_sampling}

Using $\boldsymbol{M}_\mathcal{P}^{-1}\left(n = 3\right)$, joint space samples can be generated that fulfill the displacement constraint \eqref{eq:sum_rho}.
To compute
\begin{align}
    \begin{bmatrix}
        \rho_1^{\left(k\right)} \\[0.25em]
        \rho_2^{\left(k\right)} \\[0.25em]
        \rho_3^{\left(k\right)}
    \end{bmatrix}
    =
    \begin{bmatrix} 
        1 & 0 \\[0.25em]
        -1/2 & \sqrt{3}/2 \\[0.25em]
        -1/2 & -\sqrt{3}/2
    \end{bmatrix}
    \begin{bmatrix}
        \rho_{\text{Re}, \mathcal{U}}^{\left(k\right)}\\[0.25em]
        \rho_{\text{Im}, \mathcal{U}}^{\left(k\right)}
    \end{bmatrix},
    \label{eq:sampling_MP}
\end{align}
three sampling methods for $\rho_{\text{Re}, \mathcal{U}}^{\left(k\right)}$ and $\rho_{\text{Im}, \mathcal{U}}^{\left(k\right)}$ are used.

We exploit the polar form of $\rho_\text{Re}$ and $\rho_\text{Im}$ \eqref{eq:rho_polar_form}.
Therefore, we can sample the angle
\begin{align}
    \theta_\mathcal{U} = 2\pi\,\mathcal{U}\left[0; 1\right]
    \label{eq:theta_U}
\end{align}
and amplitude $L_\mathcal{U}$ independently of each other.
We obtain
\begin{align}
    \begin{bmatrix}
        \rho_{\text{Re}, \mathcal{U}}^{\left(k\right)}\\[0.25em]
        \rho_{\text{Im}, \mathcal{U}}^{\left(k\right)}
    \end{bmatrix}
    = 
    \begin{bmatrix} 
        L_\mathcal{U}\cos\left(\theta_\mathcal{U}\right)\\[0.25em]
        L_\mathcal{U}\sin\left(\theta_\mathcal{U}\right)
    \end{bmatrix}.
    \label{eq:sampling_rhoReIm}
\end{align}
Since \eqref{eq:sampling_MP} with \eqref{eq:sampling_rhoReIm} always generated a valid $k\textsuperscript{th}$ sample, the direct sampling method can be vectorized by stacking all sampled values into one matrix.
In this case, \eqref{eq:sampling_MP} extends to
\begin{align}
    \begin{bmatrix}
        \rho_1^{\left(1\right)} & \cdots & \rho_1^{\left(k\right)} \\[0.25em]
        \vdots  & \ddots & \vdots\\[0.25em]
        \rho_n^{\left(1\right)} & \cdots & \rho_n^{\left(k\right)}
    \end{bmatrix}
    =
    \boldsymbol{M}_\mathcal{P}^{-1}
    \begin{bmatrix}
        \rho_{\text{Re}, \mathcal{U}}^{\left(1\right)} & \cdots & \rho_{\text{Re}, \mathcal{U}}^{\left(k\right)}\\[0.25em]
        \rho_{\text{Im}, \mathcal{U}}^{\left(1\right)} & \cdots & \rho_{\text{Im}, \mathcal{U}}^{\left(k\right)}
    \end{bmatrix}
    \label{eq:sampling_rho}
\end{align}
for $k$ samples.
In the following, we present three sub-types of direct sampling (referred to as (c) - (e) below), each differing in how $L_\mathcal{U}$ is sampled.

(c) The first method is straightforward and generates random values along a line:
\begin{align}
    L_\mathcal{U}
    = 
    \rho_\text{min} + \left(\rho_\text{max} - \rho_\text{min}\right)\mathcal{U}\left[0; 1\right].
    \label{eq:sampling_LU_line}
\end{align}
This approach works well for line sampling, but when applied to sampling on a disk \eqref{eq:sampling_rhoReIm}, it results in points that are more concentrated towards the center of the disk \citep{MathWorld_DiskPointPicking, GrassmannSenykBurgner-Kahrs_ICRA_2024}.

(d) To avoid this concentration effect, we employ a modified sampling method:
\begin{align}
    L_\mathcal{U}
    = 
    \rho_\text{max}\,\sqrt{\mathcal{U}\left[0; 1\right]}.
    \label{eq:sampling_LU_disk}
\end{align}
This adjustment ensures that points are uniformly distributed across the entire area of a disk with radius$\rho_\text{max}$.

(e) However, when considering joint limits, \eqref{eq:sampling_rhoReIm} actually generates points over an annulus, not a full disk. This motivates a third method, which directly samples random points within the annulus:
\begin{align}
    L_\mathcal{U}
    = 
    \sqrt{\left(\rho_\text{min}^2 + \left(\rho_\text{max}^2 - \rho_\text{min}^2\right)\mathcal{U}\left[0; 1\right]\right)}.
    \label{eq:sampling_LU_annulus}
\end{align}
Here, the inner and outer radii of the annulus are $\rho_\text{min}$ and $\rho_\text{max}$, respectively, with the condition $\rho_\text{max} > \rho_\text{min} > 0$. This method ensures uniform sampling over the entire annular region.

\subsection{Evaluation and Results}

\begin{figure}
    \centering
    \includegraphics[width=0.65\columnwidth]{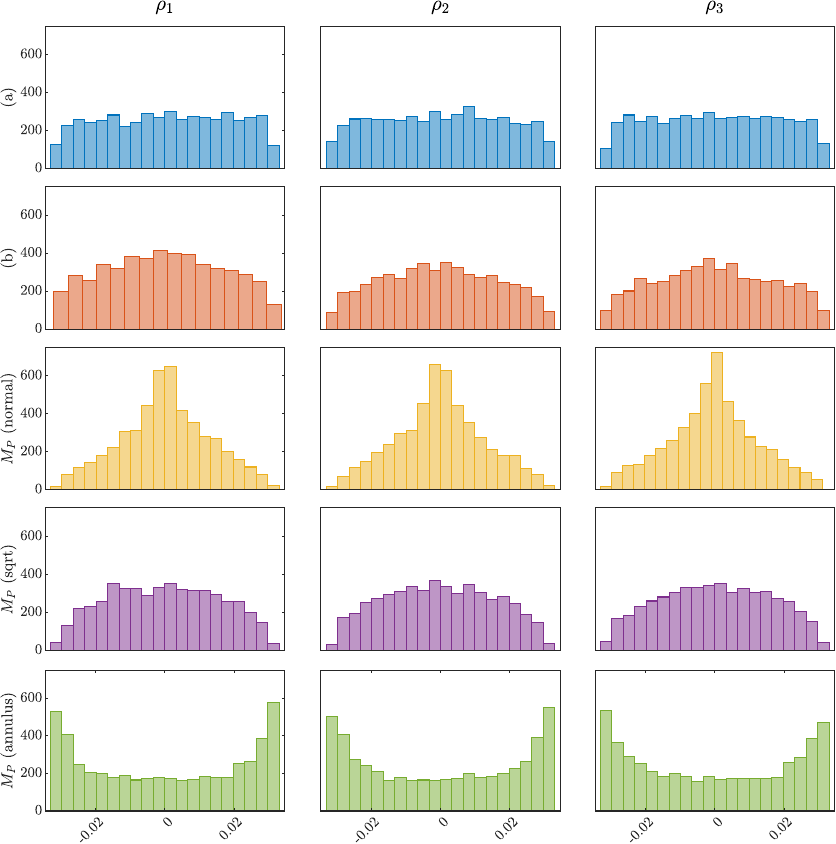}
    \caption{
        Distributions for different sampling methods.
        The columns refer to the respective sampled displacements $\rho_i$, whereas the row refers to the used sampling method. 
        The rejection sampling methods (a) and (b) are described in Sec.~\ref{sec:rejection_sampling_via_branching}. 
        The direct sampling methods using $M_\mathcal{P}$ are described in Sec.~\ref{sec:direct_sampling}, where $M_\mathcal{P}$(normal), $M_\mathcal{P}$(sqrt), and $M_\mathcal{P}$(annulus) refer to direct sampling (c), (d), and (e), respectively. 
        Table~\ref{tab:sampling_comparision} presents further details.
    }
    \label{fig:distribution}
\end{figure}

To aggregate the values listed in Table~\ref{tab:sampling_comparision} and the distributions shown Figure~\ref{fig:distribution}, a sample population of \num{5000} samples is collected by running each method five times for \num{1000} samples.
While a cloud-based \textsc{Matlab} service is used to generate the data, the relative increase of wall-clock time and number of iterations is relevant for an implementation-independent comparison.

Na\"ive sampling methods (a) and (b) employ branching and resampling, leading to increased wall-clock time compared to direct methods. 
The comparison of floating-point numbers introduces a dependency on the threshold $\epsilon$, \textit{i.e.}, a smaller $\epsilon$ results in more iterations, directly impacting wall-clock time. 
Although this is an obvious fact, it is still worth mentioning, as the value of $\epsilon$ greatly influences resampling.
According to Table~\ref{tab:sampling_comparision}, the average resampling counts are \num{826,440} for (a) and \num{344} for (b), triggered by violations of \eqref{eq:sum_rho} or $\rho_i$ boundaries.
This leads to low success rates for methods (a) and (b) being approximately $0.0012$ and $0.74$, respectively.

Conversely, sampling on the \SI{2}{dof} manifold with $\boldsymbol{M}_\mathcal{P}^{-1}$ eliminates branching, yielding four key benefits: (i) a \SI{100}{\%} success rate due to inherent consideration of the displacement constraint \eqref{eq:sum_rho}; (ii) faster execution through $2n$ multiplications for two random variables, unlike for $n$ random variables in methods (a) and (b); (iii) vectorizability, allowing all samples to be computed simultaneously, a feature unachievable with (a) and (b); and  (iv) the flexibility to alter the distribution shape, as demonstrated in examples (c)-(e) for different functions that generate the distribution.
\section{Application to Control}
\label{sec:control}

\begin{figure}[!h]
    \centering
    \includegraphics[width=.65\columnwidth]{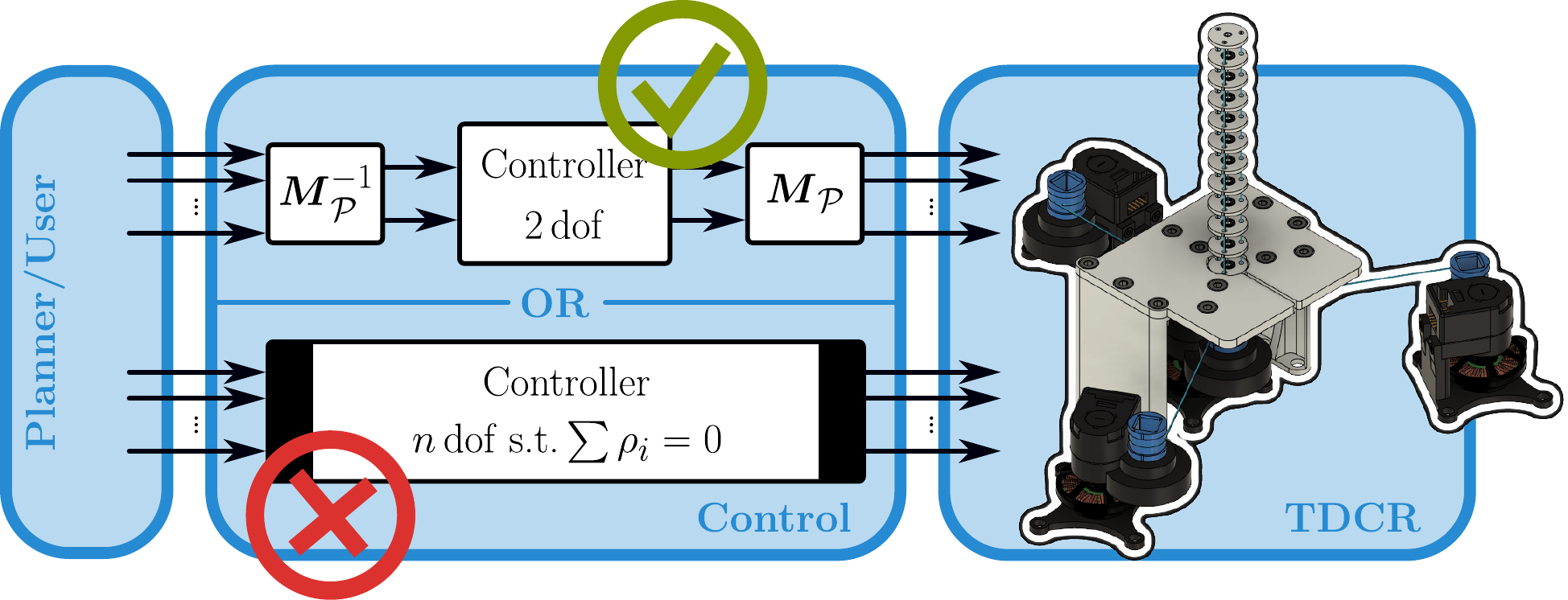}
    \caption{
        For a TDCR with $n$ tendons, controlling all $n$ tendons in joint space can lead to a non-linear controller with at least $n\,\SI{}{dof}$, which must account for displacement constraint \eqref{eq:sum_rho}. 
        The proposed Clarke transform simplifies the problem to a \SI{2}{dof} manifold within the $n\,\SI{}{dof}$  joint space, \textit{e.g.}, allowing for a simpler P controller with pre-compensation.
        }
    \label{fig:catchy_image_directClarke}
\end{figure}

In this section, we address the control of $n$ joints operating under constraint \eqref{eq:sum_rho}.
For physical continuum robots, controlling joint space displacement is essential. 
By applying the proposed Clarke transform, we can synthesize a controller on a \SI{2}{dof} manifold, reducing the need for $n$ separate PID controllers. 
This simplifies the control scheme, bypasses displacement constraints, and improves coordination of control outputs through branching. 
Additionally, it facilitates stability and convergence proofs and opens up possibilities for alternative control approaches.
Figure~\ref{fig:catchy_image_directClarke} illustrates the choice of implementation, where a TDCR is chosen as an exemplary continuum robot.

\subsection{Displacement-Constraint-Informed Controller}
\label{sec:method_control}

\begin{figure}
    \centering
    \includegraphics[width=0.65\columnwidth]{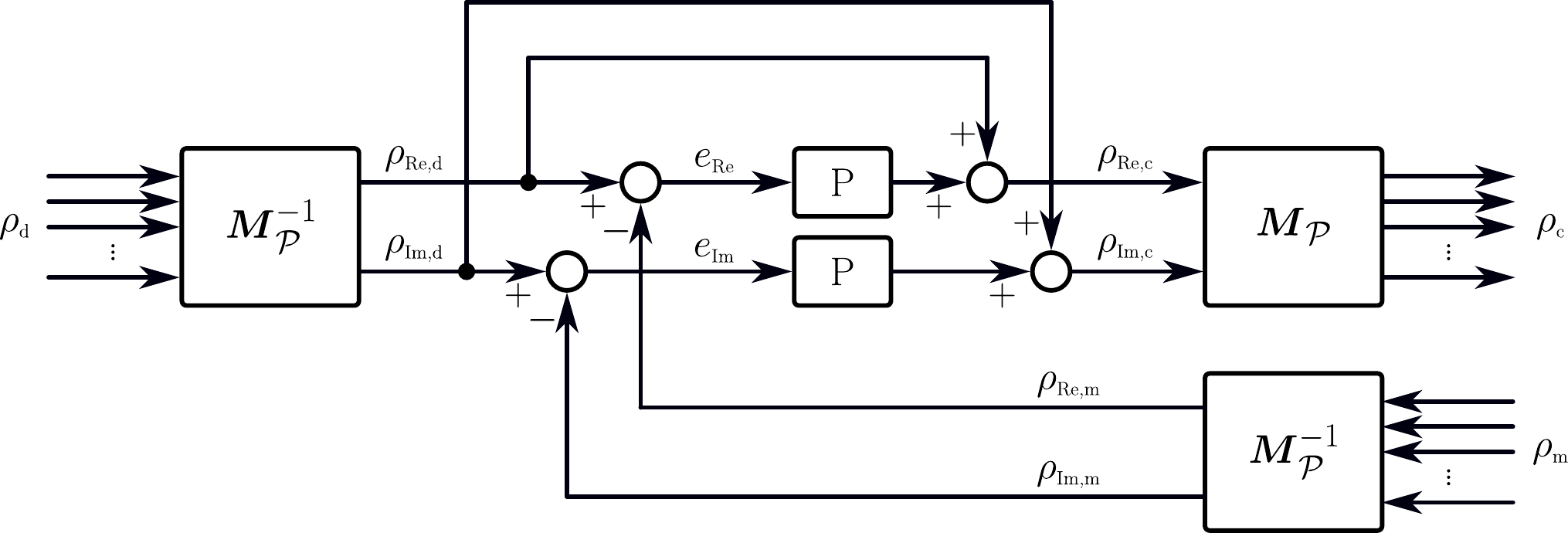}
    \caption{
    Displacement-control of $n$ joints using Clarke transform.
    Both proportional feedback controllers with precompensation are sandwiched by $\MP$ and $\MPinv$.
    }
    \label{fig:controller_precompensate}
\end{figure}

The main idea of the control scheme is to map the control problem into the manifold by using the Clark transform, which reduces the number of variables from $n$ to two.
We implement a slightly more advanced controller than a PID controller -- a proportional feedback control with precompensation as used by \cite{MorinSamson_HOR_2008}.
Figure~\ref{fig:controller_precompensate} depicts the control scheme.

As indicated in Figure~\ref{fig:controller_precompensate}, the error signal is
\begin{align}
    \begin{bmatrix}
        e_\mathrm{Re} \\ e_\mathrm{Im}
    \end{bmatrix}
    = 
    \begin{bmatrix}
        \rho_{\mathrm{Re}, \mathrm{d}} \\ \rho_{\mathrm{Im}, \mathrm{d}}
    \end{bmatrix}
    - 
    \begin{bmatrix}
        \rho_{\mathrm{Re}, \mathrm{m}} \\ \rho_{\mathrm{Im}, \mathrm{m}}
    \end{bmatrix}
    ,
    \label{eq:error_signal}
\end{align}
where the subscript $\mathrm{d}$ and $\mathrm{m}$ stand for desired and measured, respectively.
We make two remarks.
First, 
\begin{align}
    \begin{bmatrix}
        e_\mathrm{Re} \\ e_\mathrm{Im}
    \end{bmatrix}
    = 
    \boldsymbol{M}_\mathcal{P}
    \boldsymbol{\rho}_\mathrm{d}
    - 
    \boldsymbol{M}_\mathcal{P}
    \boldsymbol{\rho}_\mathrm{m}
    = 
    \boldsymbol{M}_\mathcal{P}
    \left(\boldsymbol{\rho}_\mathrm{d} - \boldsymbol{\rho}_\mathrm{m}\right)
    % \label{eq:}
    \nonumber
\end{align}
shows that the error signal \eqref{eq:error_signal} is the Clarke transform of the error between the desired displacements $\boldsymbol{\rho}_\mathrm{d}$ and the measured displacements $\boldsymbol{\rho}_\mathrm{m}$.
The second remark uses geometric insights presented in Sec.~\ref{sec:robot_dependent_mapping}, where \eqref{eq:rho_fdep_inverse} and \eqref{eq:rho_fdep} are of interest.
For the next step, either \eqref{eq:rho_fdep_inverse} is reformulated to match the Clarke coordinates, or \eqref{eq:rho_fdep} is directly inserted into the above reformulation of the error signal.
For the latter, \eqref{eq:MP_right_inverse} is necessary.
This leads to
\begin{align}
    \begin{bmatrix}
        e_\mathrm{Re} \\ e_\mathrm{Im}
    \end{bmatrix}
    = 
    ld
    \left(
    \begin{bmatrix}
        \kappa_\mathrm{d}\cos\left(\theta_\mathrm{d}\right) \\ \kappa_\mathrm{d}\sin\left(\theta_\mathrm{d}\right)
    \end{bmatrix}
    - 
    \begin{bmatrix}
        \kappa_\mathrm{m}\cos\left(\theta_\mathrm{m}\right) \\ \kappa_\mathrm{m}\sin\left(\theta_\mathrm{m}\right)
    \end{bmatrix}
    \right)
    ,
    \nonumber
\end{align}
which is equivalent to
\begin{align}
    \begin{bmatrix}
        e_\mathrm{Re} \\ e_\mathrm{Im}
    \end{bmatrix}
    = 
    ld
    \left(
    \begin{bmatrix}
        \kappa_{x, \mathrm{d}} \\ \kappa_{y, \mathrm{d}}
    \end{bmatrix}
    - 
    \begin{bmatrix}
        \kappa_{x, \mathrm{d}} \\ \kappa_{y, \mathrm{m}}
    \end{bmatrix}
    \right)
    ,
    \nonumber
\end{align}
where the error signal is linear \textit{w.r.t.} $\kappa_x$ and $\kappa_y$.

This analysis of the error signal can be seen as another advantage of the curvature-curvature representation compared to other arc space representations, see Appendix~\ref{appendix:arc_space_representation}.
That is, the control problem using the Clarke transform is a linear problem in the arc space represented by $\left(\kappa_x, \kappa_y\right)$.

\subsection{Evaluation}

To make the evaluation more concrete, we consider a TDCR, where a displacement is a tendon displacement and \eqref{eq:psi} represents the $i\textsuperscript{th}$ tendon hole location.
In particular, a TDCR with one type-0 segment, \textit{i.e.}, it has a single segment with fixed length \citep{GrassmannBurgner-Kahrs_et_al_Frontiers_2022}, and $n = 5$ tendons are considered.
The segment length is $l = \SI{0.1}{m}$, and the distance from the backbone to the tendon hole is set to $d_i = d = \SI{0.01}{m}$ for all spacer disks and tendons.

The control frequency is set to \SI{1}{kHz} resulting in a sampling time of \SI{1}{ms}, whereas the control gains are $\boldsymbol{K}_\mathrm{p} = 125$.
Each actuator is modelled as an independent first-order
proportional delay element (PT\textsubscript{1}) system, where its time constant is set to two hundred and fifty times the sampling time.
Measurement noise is imposed to $\boldsymbol{\rho}_\mathrm{m}$ modelled as additive noise drawn from a uniform distributions, \textit{i.e.},
\begin{align}
    \rhovec_\mathrm{m} \leftarrow \rhovec_\mathrm{m} + \mathcal{U}^n\left[-\epsilon, \epsilon\right],
    \label{eq:measurement_noise}
\end{align}
where $\epsilon$ is set to \SI{2.5}{mm}.
The measurement noise in \eqref{eq:measurement_noise} imitates unmodelled phenomena such as slack,  sensor noise of the encoder, or uncertainties in the actuation units.

Using the sampling method described in Sec.~\ref{sec:sampling}, five random sets of displacement vectors are generated, \textit{i.e.}, one start point, one goal point, and three via points.
Afterwards, samples are Clarke transformed via \eqref{eq:sampling_rho} using $\boldsymbol{M}_\mathcal{P}^{-1}$ for $n = 5$.
For this, we set $\theta_\text{max} = 2\pi$ and $\rho_\text{max} = d\pi$.
Note that $\rho_\text{max} = d\pi$ results in a segment forming a half-circle, where the maximal displacement is $d\pi$, \textit{cf.} \citep{Grassmann_OpenCR_2023}.

We adapted our smooth trajectory generator \citep{GrassmannBurgner-Kahrs_RAL_2019} to $n$ displacements and set the kinematic constraints to $v_i^j = \SI{0.01\pi}{m/s}$, $a_i^j = \SI{0.1\pi}{m/s^2}$, and $d_i^j = \SI{0.1\pi}{m/s^2}$ for all $i$ and $j$.
The generated paths are used as desired displacement $\rhovec_\mathrm{d}$ for the control scheme.

\begin{figure*}
    \centering
    \includegraphics[height=4.25cm, trim={0, 20, 25, 20}, clip]{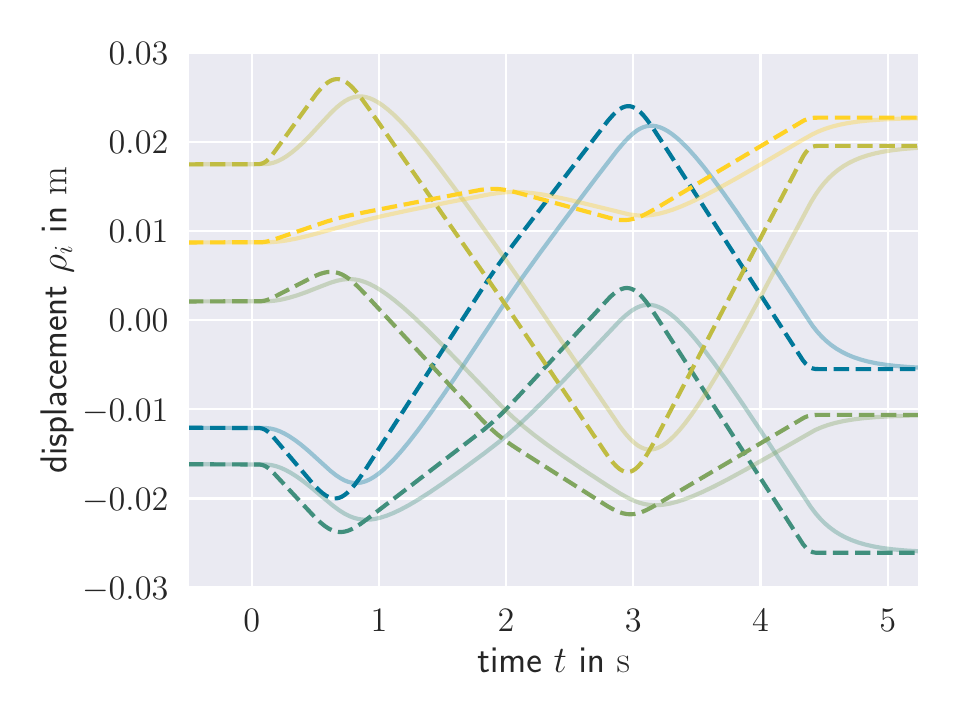}
    \hfill
    \includegraphics[height=4.25cm, trim={85, 20, 25, 20}, clip]{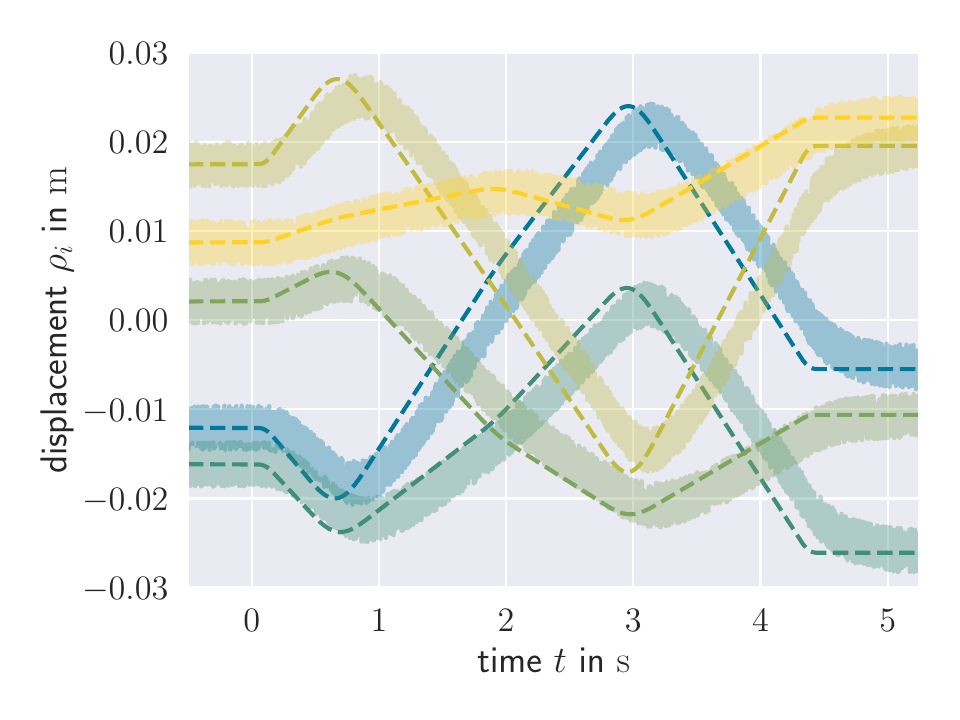}
    \hfill
    \includegraphics[height=4.25cm, trim={85, 20, 25, 20}, clip]{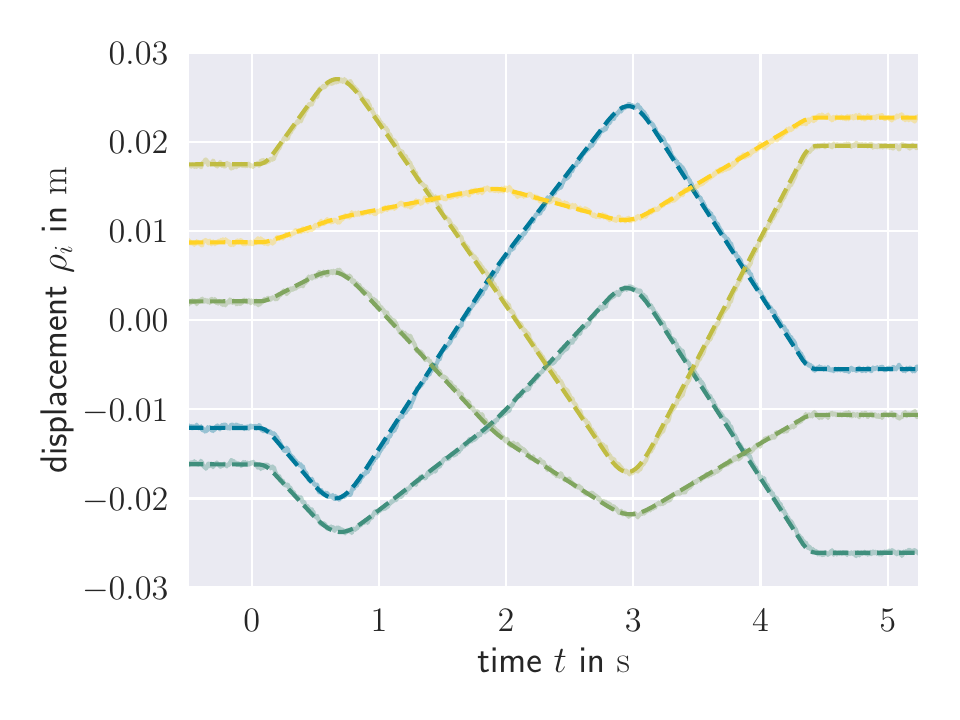}
    \caption{
        Displacement-control using pre-compensation for $n = 5$ displacements.
        (left) Desired path versus open-loop behavior of the noise-free PT1 system.
        (middle) Desired path versus measure displacement with noise.
        Due to the high frequency, the measurement noise appears to be a band.
        (right) Desired path versus closed-loop behavior.
        As can be seen, the controller can follow the desired path despite the relatively high noise and system delay.
        }
    \label{fig:results_closed_loop_advance}
\end{figure*}

Figure~\ref{fig:results_closed_loop_advance} shows the open-loop behavior with and without the noise in $\rhovec_\mathrm{m}$ and the closed-loop behavior. 
As can be seen, the controller output follows desired $\rhovec_\mathrm{d}$ despite the relatively high delay of the system and noise of $\rhovec_\mathrm{m}$.

\subsection{Noise analysis}

We look into the error contribution imposed by \eqref{eq:measurement_noise}.
Consider a noisy measurement $\rhovec_\mathrm{m}$ defined by some additive bias $\mu$ and additive zero-mean random distortions $\sigma_k$ for $k\textsuperscript{th}$ displacement.
This noisy measurement is then described by
\begin{align}
    \rhovec_\mathrm{m} = \rhovec + \mu\onevec_{n \times 1} + \sum_{k=1}^{n}\sigma_k\hotvec_{n \times 1}^{(k)}
    ,
    \nonumber
\end{align}
where $\mu\onevec$ is an offset, $\sigma_k\hotvec^{k}_{1 \times n}$ an error contribution of the $k\textsuperscript{th}$ joint, and, for the ideal case, $\rhovec$ is equal to $\rhovec_\mathrm{d}$.
Using the properties \eqref{eq:properties_vanishing_bias} and \eqref{eq:properties_hotvec}, the bias $\mu$ vanished under the Clarke transform and the transformed $\rhovec_\mathrm{m}$ results in
\begin{align}
    \MP\rhovec_\mathrm{m} 
    = 
    \rhoclarke
    +
    \sum_{k=1}^{n}\sigma_k
    \begin{bmatrix}
        \cos\left(\psi_k\right) \\
        \sin\left(\psi_k\right)
    \end{bmatrix}
    ,
    \nonumber
\end{align}
where $\rhoclarke = \MP\rhovec$.
Note that $\rhovec_\mathrm{m} \in \mathbb{R}^n$ but $\rhovec_\mathrm{m} \notin \mathbf{Q}$.

Consider the next time step, where the above expression is transformed back to the joint space.
After applying \eqref{eq:inverse_mapping}, the sharp $\hotvec_{n \times 1}^{(k)}$ is now spread out to all other displacements, see property \eqref{eq:properties_circle_manifold}.
Furthermore, the peak at the $k\textsuperscript{th}$ entry has the same value, before, \textit{i.e.}, $\sigma_k\hotvec^{k}_{n \times 1}$, and after, \textit{i.e.}, $\sigma_k\MPinv\MP\hotvec^{k}_{n \times 1}$, the transformation, recall Figure~\ref{fig:unit_circle_on_manifold} for visual aid.
As a consequence, the magnitude of error is amplified by $\sqrt{n/2}$.
We can derived
\begin{align}
    \left(\MPinv\MP\sigma_k\hotvec_{n \times 1}^{(k)}\right)\transpose\left(\MPinv\MP\sigma_k\hotvec_{n \times 1}^{(k)}\right)
    &= \dfrac{n\sigma_k^2}{2}
    ,
\end{align}
where, in this order, \eqref{eq:MP_transpose_relation}, \eqref{eq:MP_right_inverse}, \eqref{eq:properties_hotvec}, and a trigonometric identity are used to show this.
In contrast, $\left(\sigma_k\hotvec_{n \times 1}^{(k)}\right)\transpose\sigma_k\hotvec_{n \times 1}^{(k)} = \sigma_k^2\hotvec_{1 \times n}^{(k)}\hotvec_{n \times 1}^{(k)} = \sigma_k^2$.
Note that, due to property \eqref{eq:MP_Toeplitz_idempotent}, the peak is not further amplified over the time, \textit{i.e.}, by repeated multiplication with $\MPinv\MP$.
Furthermore, note that $\MPinv\MP\rhovec_\mathrm{m} \in \mathbf{Q}$ but $\rhovec_\mathrm{m} \notin \mathbf{Q}$.
This analysis shows that one faulty joint can have ripple effects on the other joint.
\section{Discussion}
\label{sec:discussion}

We derived a generalized Clarke transformation matrix for arbitrary $n$, extending existing methods such as \cite{Janaszek_PIE_2016}, \cite{Willems_TOE_1969}, and \cite{RockhillLipo_IEMDC_2015}. 
This transformation is amplitude-invariant, which enables direct comparison between the magnitude of the Clarke coordinates and the bending achieved, regardless of the number of actuators.

Our work is based on the displacement constraint \eqref{eq:sum_rho} which entangles the displacements of all joints in continuum robots. 
Using the Clarke transform, we reduce this complex joint space to a \SI{2}{dof} manifold, thereby disentangling the displacements. 
The Clarke coordinates have a clear physical interpretation, as they correspond to the curvature and bending parameters commonly used in arc parameter based models for continuum robots. 
By linking the Clarke coordinates to these arc parameters, we create a unified framework that can incorporate conventional arc representations and provide a solution to the previously unknown forward robot-dependent mapping for arbitrary $n$.

The use of the Clarke transform enables a closed-form solution for continuum robot kinematics. 
The single-segment solution enables solutions of the multi-segment problem if the position $\boldsymbol{p}$ of each segment is known \citep{NeppalliWalker_et_al_AR_2009}.
For this case, the solution presented in Sec.~\ref{sec:kinematics} can be used to derive a closed-form solution for a multi-segment robot.
This closed-form solution eliminates branching and provides a computationally efficient framework, even for robots with multiple segments. 
The ability to compute kinematics in a straightforward, non-iterative manner opens possibilities for real-time applications and facilitates further advancements in multi-segment robot control.

Instead of dealing with $n\,\SI{}{dof}$ systems, the Clarke transform isolates two key variables, significantly simplifying the system and streamlining the control process.
For instance, for tendon-driven continuum robots, the Clarke transform simplifies the handling of actuation displacement constraints, ensuring smooth control without the need for complex coordination between tendons. 
This is especially useful for robots with fully constrained tendon paths, where the displacement constraints must be rigorously satisfied. 
The transform ensures that the system remains stable and has the potential to avoid issues such as tendon slack or overextension.

Various state parameterizations, such as those introduced by \cite{DianGuo_et_al_Access_2022} and \cite{DellaSantinaBicchiRus_RAL_2020}, can be unified through the Clarke transform. 
By generalizing these methods for $n \geq 3$ displacements, the Clarke transform provides a consistent approach that applies across different types of continuum robots, including tendon-driven, pneumatic, and soft robots. 
This unification reduces the need for specialized solutions and enables easier transfer of knowledge across different robot morphologies and control systems.

The Clarke transform is not limited to continuum robots; its principles apply to a range of robotic systems and mathematical frameworks. 
By facilitating knowledge transfer between fields such as soft robotics, cable-driven robots, and continuum robots, the Clarke transform promotes interdisciplinary collaboration. 
This flexibility makes it a valuable tool for researchers working across different domains, allowing for a more holistic approach to robotic control, design, and modeling.
\section{Future Work}
\label{sec:futurework}

Building upon the generalized Clarke transformation matrix for arbitrary $n$, the potential benefits of utilizing a larger number of displacement-actuated joints per continuum robot segment are considered. 
Increasing $n$ significantly (\textit{i.e.}, $n\!\gg\!3$) offers several advantages: (i) enhanced manipulability, particularly between bending directions that coincide with joint locations \eqref{eq:psi} and backbone positions; (ii) improved force absorption and delivery due to distributed nature of the actuation forces; and (iii) increased safety through actuation redundancy. 
These advantages are particularly desirable in medical applications \citep{Burgner-KahrsRuckerChoset_TRO_2015, DupontRucker_et_al_JPROC_2022} and industrial settings \citep{DongKell_et_al_JMP_2019, RussoAxinte_et_al_AIS_2023}. 
Therefore, exploiting a higher number of displacement-actuated joints per segment could be highly beneficial. 
It is hypothesized that this approach may lead to increased load capacity, better shape conformation, enhanced stability, and variable stiffness.

For multi-segment continuum robots with displacement-actuated joints independent of segment sequence (\textit{e.g.}, pneumatically actuated soft robot arms), a diagonal block matrix with $\boldsymbol{M}_\mathcal{P}$ suffices. 
Similarly, TDCRs with multiple segments using identical tendons and pre-compensated displacements can be simplified. 
However, in most TDCR designs, distal tendon displacements affect proximal segments and must be compensated by subtracting distal from proximal displacements. 
Leveraging the linearity of the forward mapping \eqref{eq:forward_mapping}  and insights from Figure~\ref{fig:rho_physical_interpretation}, segment-wise Clarke coordinates can be utilized. 
This leads to an upper triangular block matrix with $\boldsymbol{M}_\mathcal{P}$ as non-zero entries and negative off-diagonal elements. 
This formulation adapts to current and future methods to multi-segment robots, allows for varying joint numbers per segment (enabling reinforced proximal segments), and benefits piece-wise constant curvature models.

Finally, studying the generalization and application of the Clarke transform to vector-valued functions of time and arc length along the robot’s backbone could be highly beneficial. 
This approach would enable the integration of widely used models based on Cosserat and Kirchhoff rod theories. Furthermore, extending the piece-wise constant curvature model to accommodate variable curvature (changing with respect to the arc length $s$) could enhance modeling accuracy. 
Adapting this framework to dynamic modeling might also lead to more computationally efficient dynamic models.
\section{Conclusions}
\label{sec:conclusion}

This work introduces the Clarke transform and generalized Clarke coordinates as fundamental tools for simplifying continuum robot analysis. 
By demonstrating their physical meaning and mathematical consistency, the transform’s utility was extended across applications in kinematics, sampling, and control. 
Each application emphasizes the simplicity, interpretability, and scalability of any number of joints. 
The Clarke transform and Clarke coordinates offer a mathematically consistent approach that can unify methods across different types of continuum robots, providing a robust foundation for future developments in continuum robot modeling and control.

\section*{Funding}
We acknowledge the support of the Natural Sciences and Engineering Research Council of Canada (NSERC), [RGPIN-2019-04846].

\appendix
\section{Intuition and Analogy}
\label{appendix:intuition}

We provide our intuition and used analogies.
While both are not relevant for the presentation of our solution, they might give guidance for future approaches as they were useful to us in establishing the necessary connections and sparking key ideas.

\subsection{Observation and Na\"ive Construction of the Manifold}
\label{appendix:observation}

\begin{figure}
\centering
    \includegraphics[width=0.55\columnwidth]{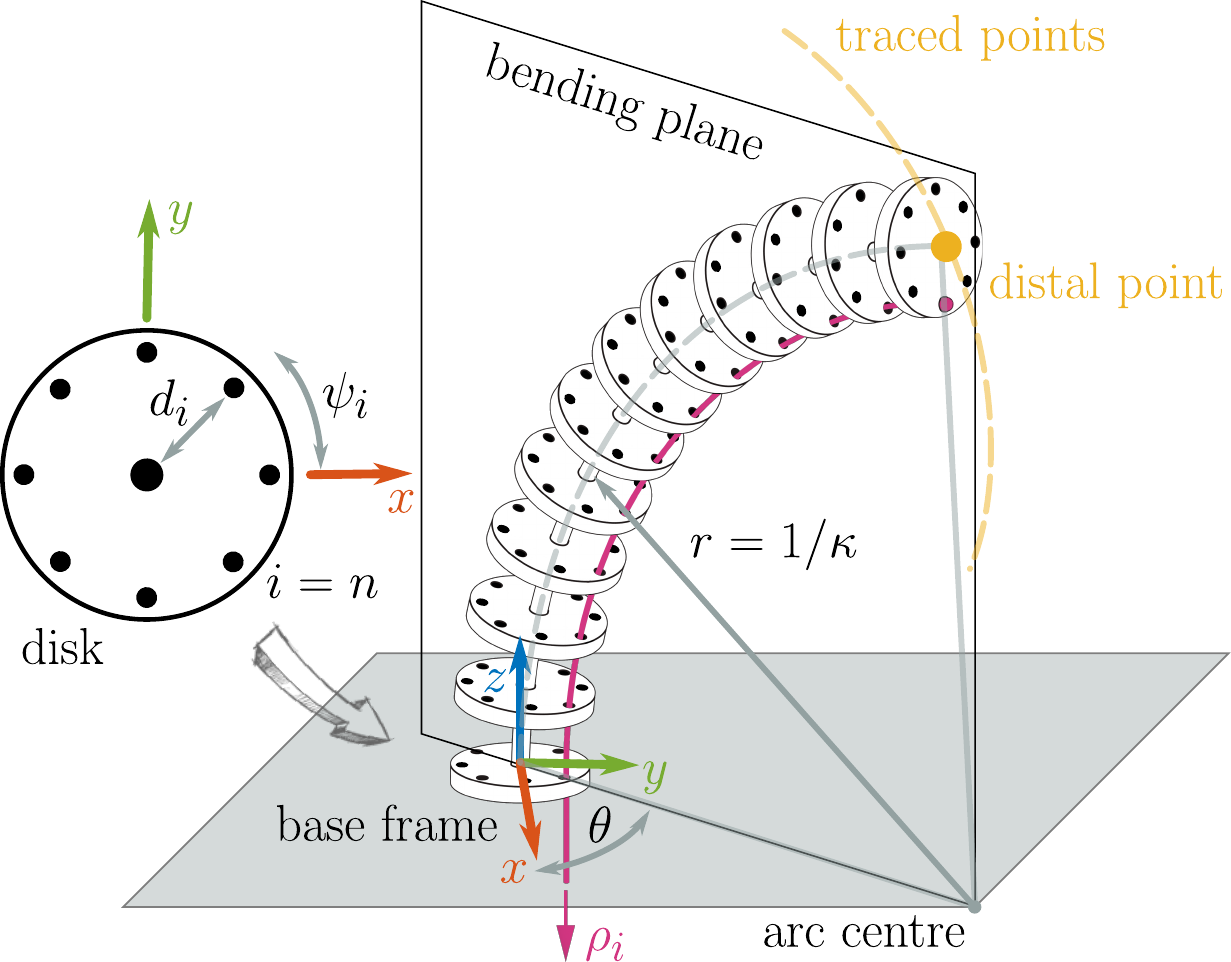}
    \caption{
    Illustration of a bend TDCR.
    It is actuated by tendon displacements denoted by $\rho_i$.
    Note that only one of the eight tendons is depicted.
    The TDCR can be fully described by a curricular arc and its arc parameters: curvature $\kappa$ and bending plane angle $\theta$.
    On the disk, the $i\textsuperscript{th}$ tendon hole location is described by $d_i$ and $\psi_i$.
    Tracing the position of the distal spacer disk for different curvatures $\kappa$, the traced points form a line that coincides with the bending plane.
    Afterward, rotating the bending plane will create all reachable points in the task space, which is a two-dimensional dome-like surface.
    }
    \label{fig:tdcr}
\end{figure}

To provide a more concrete example, let us focus on TDCR with one continuum segment, see Figure~\ref{fig:tdcr}.
Furthermore, the discussion is also restricted to $n = 3$ and $n = 4$ tendons.
It is commonly assumed that the tendons are equally distributed around the backbone.
For a segment, those tendons are constrained by
\begin{align}
	\sum_{i=1}^{n} \rho_{i} = 0,
    \nonumber
\end{align}
where $\rho_{i}$ denotes the tendon displacement of the $i^\text{th}$ tendon.
It is commonly known that $n$ tendons map onto a dome-like surface in task space or two variables in the arc space.
Therefore, we can expect a \SI{2}{dof} manifold embedded in the joint space of $n$ tendons.
Note that, for $n = \{1, 2\}$, the dome-like surface degenerates into a line.

For three tendons, the sum of all $\rho_i$ is constrained by
\begin{align}
	\sum_{i=1}^{n=3} \rho_{i} = \rho_1 + \rho_2 + \rho_3 = 0. \nonumber
\end{align}
The resulting joint space is a manifold described by the set
\begin{align}
	\mathbf{Q}^{(3)} = \Set{ \left(\rho_1, \rho_2, \rho_3\right) \in \mathbb{R}^3 | \rho_1 + \rho_2 + \rho_3 = 0},
    \nonumber
\end{align}
where the superscript in $\mathbf{Q}^{(3)}$ indicates the number of tendons.
To resolve the displacement constraint, $\rho_1$ and $\rho_2$ are actuated and $\rho_3$ is set to $\rho_3 = -\rho_1 - \rho_2$.
Depending on the employed heuristic, the last variable can be expressed by the other reminding tendon displacements using cyclic permutation, \textit{i.e.}, $\rho_1 = -\rho_2 - \rho_3$ or $\rho_2 = -\rho_3 - \rho_1$.
Therefore, this commonly used simplification comes with caveats, where the practical implementation relies on heuristics with branching. \textit{i.e.}, selecting two suitable tendon displacements from three possibilities. 

For four tendons, a convenient simplification can be found.
First of all, the displacement constraint expenses to
\begin{align}
	\sum_{i=1}^{n=4} \rho_{i} = \rho_1 + \rho_2 + \rho_3 + \rho_4 = 0. 
    \nonumber
\end{align}
However, this displacement constraint only provides one equation, which is not sufficient to describe the sought-after \SI{2}{dof} manifold.
Fortunately, by exploiting symmetries observed in physical prototypes, the displacement constraint simplifies to two constraints, which are given by
\begin{align}
	\rho_1 + \rho_3 = 0 \quad\text{and}\quad \rho_2 + \rho_4 = 0.
    \nonumber
\end{align}
As a beneficial side-effect, this choice of constraints prevents geometric unfeasible actuation such as $\rho_1 = -\rho_2$ and $\rho_3 = - \rho_4$, which is possible by only considering the sum of all tendon displacement.
Having found the two equations, we can now define the joint space $\mathbf{Q}^{(4)}$ being the set of possible ordered quadruples defined as
\begin{align}
    \small
    \mathbf{Q}^{(4)} = 
	\Set{\!\left(\rho_1, \rho_2, \rho_3, \rho_4\right)\in\mathbb{R}^4 | \rho_1 + \rho_3 = 0 \wedge \rho_2 + \rho_4 = 0\!}. 
    % \label{eq:manifold_4_tendons}
    \nonumber
\end{align}

To recap, this formulation gives a correct number of equations to constrain a four-dimensional space to a \SI{2}{dof} manifold.
Note that a formulation using the sum of all tendon displacements will not provide the correct number of equations and set definitions.
Furthermore, the two simpler constraints directly suggest a mode of actuation -- the differential actuation.

In general, tricks utilized for $\mathbf{Q}^{(3)}$ and $\mathbf{Q}^{(4)}$ cannot work for an arbitrary high-dimensional joint space $\mathbf{Q}$.
One notable exception is $n = 6$, where three pairs form a differential actuation.
A rather unsuitable approach is to use branching or sampling, considering displacement constraints for a high number of tendons.

\subsection{Kirchhoff's Current Law and Motor Control}
\label{appendix:kirchhoff}

Our intuition to look into the Clarke transformation matrix is based on our observations and an analogy gained through our work on the hardware \citep{GrassmannBurgner-Kahrs_et_al_Frontiers_2024} of the \href{https://www.opencontinuumrobotics.com/}{Open Continuum Robotics Project}.

Upon examining, the displacement constraint, \textit{i.e.}, $\sum_{i=1}^{n} \rho_i = 0$, appears analogous to Kirchhoff's current law, which posits that the net electric current through a junction is zero.
Interpreting this through effort and flow \citep{GawthropBevan_CSM_2007}, electric current (the rate of charge change) represents flow. 
Furthermore, by dividing the constraint by a small time step $\Delta t$, \textit{i.e.}, $(1/\Delta t) \sum_{i=1}^{n} \rho_i = 0$, it approaches a velocity constraint in the limit, expressed as $\sum_{i=1}^{n} \dot{\rho}_i = 0$, where $\dot{\rho}_i$ denotes flow.
Note that $\rho_i$ is a displacement and, therefore, the change between the straight configuration and the bend current configuration of a continuum robot.
Hence, Kirchhoff's law aligns with tendon velocity constraints. 
This analogy further extends to the principle that the total charge in electrical circuits remains constant, analogous to the premise of $\sum_{i=1}^{n} \rho_i = 0$.
Although this comparison is not strictly rigorous, it sufficiently motivates further exploration of this analogy.

Furthermore, Kirchhoff's current law is essential for controlling a brushless electric motor with three phases. 
Field-oriented Control (FOC), also known as Vector Control, is a pivotal method for these motors that avoid mechanical commutators. 
FOC employs the Clarke transformation (also known as alpha-beta transformation), a linear mapping represented by a constant matrix, aiming to reduce three electric current values to two. 
The used Clarke transformation matrix $\boldsymbol{M}_\text{Clarke}$ transforms three currents in a set of three transformed values, where one value is always zero if the system is balanced.
Therefore, the Clarke transformation matrix is a good starting point to find the sought-after reduction of three values to two values.
However, it is limited to $n = 3$.
Nevertheless, the core idea of this article is to exploit and extend the Clarke transformation matrix to project $n$ phases, in our case, $n$ displacements, into two variables.
\section{Derivation of Robot-independent Mapping}
\label{appendix:robot_independent_mapping}

\begin{figure}
    \centering
    \includegraphics[width=.55\columnwidth]{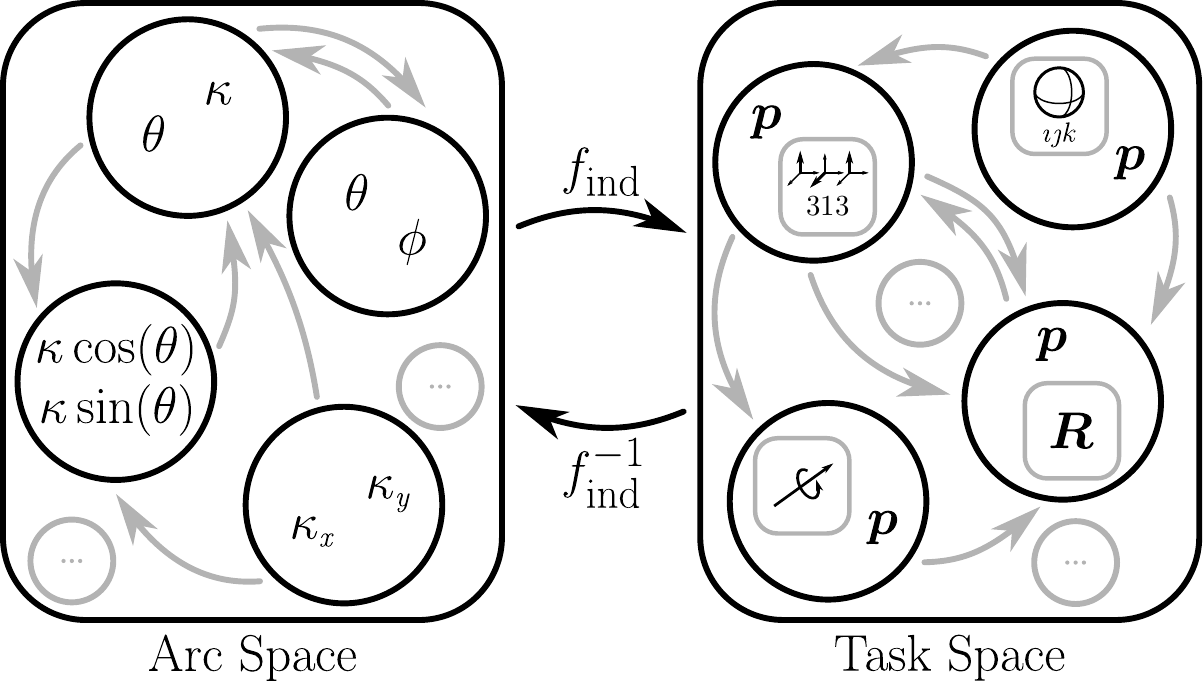}
    \caption{
        Various representations to consider for robot-independent mapping.
        The robot-independent mapping is denoted by $f_\text{ind}$, and its inverse mapping is denoted by $f_\text{ind}^{-1}$.
        Various task space representation exists for the task space.
        The arc space can be parameterized using different representations.
    }
    \label{fig:robot-dependent-mapping}
\end{figure}

We revisit the robot-independent mappings between the arc space and task space.
First, we point out that various representation exists for the task space as well as arc space as illustrated in Figure~\ref{fig:robot-dependent-mapping}. 
Second, the forward robot-independent mapping $f_\text{ind}$ is stated.
Afterward, its inverse mapping, \textit{i.e.}, inverse robot-independent mapping $f_\text{ind}^{-1}$, is derived.

\subsection{Various Representations}

We consider two different representations of a single type-0 segment \citep{GrassmannBurgner-Kahrs_et_al_Frontiers_2022} with constant curvature characteristics as depicted in Figure~\ref{fig:tdcr}, \textit{i.e.}, curvature-curvature representation and curvature-angle representation.
The mapping between the curvature-curvature representation to curvature-angle representation, \textit{i.e.}, 
\begin{align}
	 \kappa = \sqrt{\kappa_x^2 + \kappa_y^2} \quad\text{and}
  \label{eq:kappa_from_ccr}
  \\
  \quad \theta = \arctantwo\left(\kappa_y, \kappa_x\right)
  \label{eq:theta_from_ccr}
\end{align}
and its inverse mapping given by \eqref{eq:kappa_xy_from_car} are useful for the derivation.
Figure~\ref{fig:robot-dependent-mapping} shows and Table~\ref{tab:arc_space_representation} lists various representations within the arc space.
A discussion on all four representations is given in Appendix~\ref{appendix:arc_space_representation}.
Note that, for all representations, segment length $l > 0$ is constant and, therefore, it is rather a design parameter than a variable in the arc space.

For the task space, the rotation matrix
\begin{align}
    \Rm = 
    \begin{bmatrix}
        r_{11} & r_{12} & r_{13}\\[0.125em]
        r_{21} & r_{22} & r_{23}\\[0.125em]
        r_{31} & r_{32} & r_{33}
    \end{bmatrix}
    \label{eq:R_measured}
\end{align}
provides information about the tip orientation, whereas
\begin{align}
    \pv = 
    \begin{bmatrix}
        p_x & p_y & p_z 
    \end{bmatrix}
    \transpose
    \label{eq:p_measured}
\end{align}
describes the tip position providing translational information.

\subsection{Forward Robot-Independent Mapping}

Using the curvature-angle representation, the classic forward robot-independent mapping $f_\text{ind}$ can be formulated.
The mapping is given by the position vector
\begin{align}
    \pv_\mathrm{ind} &= 
    \begin{bmatrix}
        \cos\left(\theta\right)\left(1 - \cos\left(\kappa l\right)\right) / \kappa\\
        \sin\left(\theta\right)\left(1 - \cos\left(\kappa l\right)\right) / \kappa\\
        \sin\left(\kappa l\right) / \kappa
    \end{bmatrix}
    \label{eq:p_indep}
\end{align}
and the rotation matrix
\begin{align}
    \small
    \Rm_\mathrm{ind} =
    \begin{bmatrix}
        \cos\left(\theta\right)\cos\left(\kappa l\right) & -\sin\left(\theta\right) & \cos\left(\theta\right)\sin\left(\kappa l\right)\\[0.25em]
        \sin\left(\theta\right)\cos\left(\kappa l\right) & \cos\left(\theta\right) & \sin\left(\theta\right)\sin\left(\kappa l\right) \\[0.25em]
        -\sin\left(\kappa l\right) & 0 & \cos\left(\kappa l\right)
    \end{bmatrix}
    .
    \label{eq:R_indep}
\end{align}
Note that \eqref{eq:R_indep} describes a variation to represent the tip orientation.
Based on Bishop's frame, \cite{WebsterJones_IJRR_2010} presents an alternative tip orientation.

In the special case of no bending, \textit{i.e.}, $\kappa = 0$, we make the reasonable assumption that \eqref{eq:R_indep} is an identity matrix $\boldsymbol{I}_{3\times3}$.
This implies $r_{22} = \cos\left(\theta\right) = 1$ and, therefore, we define
\begin{align} 
    \theta\left(\kappa = 0\right) = 0
    \quad\text{and}\quad
    \Rm_\mathrm{ind}\left(\theta = 0, \kappa = 0\right) = 
    \boldsymbol{I}_{3\times3}
    \label{eq:R_indep_straight}
\end{align}
for a straight segment.
Furthermore, we can set \eqref{eq:p_indep} to
\begin{align}
    \pv_\mathrm{ind}\left(\theta = 0, \kappa = 0\right) = 
    \begin{bmatrix}
        0 & 0 & l 
    \end{bmatrix}
    \transpose
    ,
    \label{eq:p_indep_straight}
\end{align}
see Sec.~\ref{sec:singularities} for a derivation.
Note that, for a straight configuration, \eqref{eq:kappa_from_ccr} implies $\kappa_x = \kappa_y = 0$.
However, \eqref{eq:theta_from_ccr} is undefined for $\kappa_x = \kappa_y = 0$.

\subsection{Inverse Robot-Independent Mapping}
\label{sec:inverse_robot_independent_mapping}

We derive the solution for different sources of information; position from $E\!\left(3\right)$, orientation from $SO\!\left(3\right)$, or both, \textit{i.e.}, pose from $SE\!\left(3\right)$.
The solution method outlined in this section is straightforward and mentioned in the first textbook in robotics by \cite{Paul_book_1981}.
Furthermore, we assume $\kappa \not= 0$, otherwise the solution can be directly obtained using \eqref{eq:p_indep_straight} and \eqref{eq:R_indep_straight}.
For practical reasons, the case for $p_z \leq 0$ and $p_x = p_y = p_z = 0$ are prohibited.
All derived solutions are listed in Table~\ref{tab:solution_find_inv}.

\paragraph{Tip Position}

Given \eqref{eq:p_measured} and \eqref{eq:p_indep}, we can use a trigonometric identity, \textit{i.e.}, $1 - \cos\left(\kappa l\right) = \tan\left(\kappa l/2\right)\sin\left(\kappa l\right)$.
We can restate the relation, which gives us
\begin{align}
	 p_x &= \cos\left(\theta\right)\tan\left(\kappa l / 2\right)\sin\left(\kappa l\right) / \kappa, 
  \label{eq:p_x_position}
  \\
	 p_y &= \sin\left(\theta\right)\tan\left(\kappa l / 2\right)\sin\left(\kappa l\right) / \kappa, \quad\text{and} 
  \label{eq:p_y_position}
  \\
	 p_z &= \sin\left(\kappa l\right) / \kappa. 
  \label{eq:p_z_position}
\end{align}

For the representation curvature-angle representation, $\kappa$ and $\theta$ need to be found by using \eqref{eq:p_x_position} and \eqref{eq:p_y_position}.
The angle $\theta$ can be recovered by
\begin{align}
	 \theta = \arctantwo\left(p_y, p_x \right)
  .
  \label{eq:theta_position}
\end{align}
To find $\kappa$, we first sum the squares of \eqref{eq:p_x_position} and \eqref{eq:p_y_position}, where \eqref{eq:p_z_position} is inserted as well. 
This leads to $p_x^2 + p_y^2 = \tan^2\left(\kappa l / 2\right) p_z^2$, where $\cos^2\left(\theta\right) + \sin^2\left(\theta\right) = 1$.
Afterward,
\begin{align}
    \tan\left(\kappa l / 2\right) &=  \sqrt{\left(p_x^2 + p_y^2\right)/p_z^2}
    \label{eq:intermediate_step_tan}
\end{align}
can be derived.
Next, a double angle identity is adapted, \textit{i.e.}, $\sin\left(\kappa l\right) = \left(2\tan\left(\kappa l/2\right)\right)/\left(1 + \tan^2\left(\kappa l/2\right)\right)$, and used as a substitution applied to \eqref{eq:p_z_position}. 
This leads to
\begin{align}
    p_z & = 2\tan\left(\kappa l / 2\right)/\left(\kappa\left(1 + \tan^2\left(\kappa l / 2\right)\right)\right).
    \label{eq:intermediate_step_pz}
\end{align}
Afterward, \eqref{eq:intermediate_step_tan} is substituted in \eqref{eq:intermediate_step_pz} to derive
\begin{align}
	 \kappa &= 2\sqrt{p_x^2 + p_y^2}\ /\left(p_x^2 + p_y^2 + p_z^2\right).
  \label{eq:kappa_position}
\end{align}

For the curvature-curvature representation, one realizes that the arguments of \eqref{eq:theta_from_ccr} and \eqref{eq:theta_position} expressed as $\arctan$ functions can be set equal.
This leads to the ratios
\begin{align}
  \kappa_y / \kappa_x = p_y / p_x \quad\text{and}\quad \kappa_x / \kappa_y = p_x / p_y . 
  \nonumber
\end{align}
To find $\kappa_x$ and $\kappa_y$, we first equate $\theta = \arctantwo\left(\kappa_y, \kappa_x\right)$ and \eqref{eq:kappa_position}.
Second, the above ratio is rearranged to $\kappa_x$ and $\kappa_y$.
Third, the result of the second step is inserted into the result of the first step.
After rearranging, we can find
\begin{align}
	 \kappa_x = \dfrac{2p_x}{p_x^2 + p_y^2 + p_z^2} 
  \quad\text{and}\quad
  \kappa_y = \dfrac{2p_y}{p_x^2 + p_y^2 + p_z^2}
  .
  \label{eq:kappa_xy_position}
\end{align}
As a sanity check, by considering \eqref{eq:kappa_from_ccr}, one can derive \eqref{eq:kappa_position} by using \eqref{eq:kappa_xy_position}.

\paragraph{Tip Orientation}

Now, only a measured tip orientation \eqref{eq:R_measured} is provided.
Starting with the curvature-angle representation, the entries $r_{12}$, $r_{22}$, $r_{31}$, and $r_{33}$ of \eqref{eq:R_indep} are immediately noticeable and can be used to derive
\begin{align}
	 \theta &= \arctantwo\left(-r_{12}, r_{22}\right) \quad\text{and} \label{eq:theta_orientation}\\
  \kappa &= \left(1 / l \right)\arctantwo\left(-r_{31}, r_{33}\right). \label{eq:kappa_orientation}
\end{align}
Note that $\theta = \arctantwo\left(r_{11}^2 + r_{13}^2, r_{21}^2 + r_{23}^2\right)$ and $\kappa l = \arctantwo\left(r_{13}^2 + r_{23}^2, r_{11}^2 + r_{21}^2\right)$ can be derived.
However, both cannot cover all four quadrants of $\theta$ and $\kappa l = \phi$, while \eqref{eq:theta_orientation} and \eqref{eq:kappa_orientation} can.
Further note that $\kappa l = \arccos\left(r_{33}\right)$ sufficient to met the requirement of $\kappa \in \mathbb{R}_{\geq 0}$, Appendix~\ref{appendix:arc_space_representation}.

For the curvature-curvature representation, we use a similar approach as shown for \eqref{eq:kappa_xy_position} and establish the ratios
\begin{align}
  {\kappa_y} / {\kappa_x} = \left(-r_{12}\right) / {r_{22}} \quad\text{and}\quad {\kappa_x} / {\kappa_y} = {r_{22}} / \left(-r_{12}\right)
  \nonumber
\end{align}
by using \eqref{eq:theta_from_ccr} and \eqref{eq:theta_orientation}.
Following similar steps as described for \eqref{eq:kappa_xy_position}, and utilizing \eqref{eq:kappa_from_ccr} and \eqref{eq:theta_orientation}, the curvatures
\begin{align}
	 \kappa_x &= \left( r_{22} / l \right)\arctantwo\left(-r_{31}, r_{33}\right)
  \quad\text{and}
  \label{eq:kappa_x_orientation}
  \\
  \kappa_y &= \left( r_{12} / l \right)\arctantwo\left(-r_{31}, r_{33}\right)
  \label{eq:kappa_y_orientation}
\end{align}
can be found.
Considering \eqref{eq:R_indep}, we utilize the fact that $r_{12}^2 + r_{22}^2 = \sin^2\left(\theta\right) + \cos^2\left(\theta\right) = 1$ for \eqref{eq:kappa_x_orientation} and \eqref{eq:kappa_y_orientation}.
As a sanity check, substituting \eqref{eq:kappa_x_orientation} and \eqref{eq:kappa_y_orientation} in \eqref{eq:kappa_from_ccr} results in \eqref{eq:kappa_orientation}.

To solve the inverse robot-independent mapping, \eqref{eq:kappa_orientation} and \eqref{eq:theta_orientation}, or \eqref{eq:kappa_x_orientation} and \eqref{eq:kappa_y_orientation} can be used.
Note that \eqref{eq:R_indep} does not have an entry with an isolated $\kappa$ or $l$.
In other words, $\kappa$ and $l$ are always given as a product.
Further note that \eqref{eq:kappa_orientation}, \eqref{eq:kappa_x_orientation}, and \eqref{eq:kappa_y_orientation} include $l$.
Therefore, tip orientation is given as a rotation matrix \eqref{eq:R_measured} does not provide enough information to solve for $l$.

\paragraph{Tip Pose}

For the curvature-angle representation, we identify $-r_{31} = \sin\left(\theta\right)$ in \eqref{eq:R_measured}, and $\kappa p_z = \sin\left(\theta\right)$ in \eqref{eq:p_z_position} leading to
\begin{align}
	 \kappa &= -r_{31} / p_z. 
  \label{eq:kappa_pose}
\end{align}
For $\theta$, no simpler expressions as \eqref{eq:theta_position} and \eqref{eq:theta_orientation} can be found.

For the curvature-curvature representation, we utilize \eqref{eq:kappa_x_orientation} and \eqref{eq:kappa_y_orientation} as the starting point.
Now, \eqref{eq:kappa_orientation} and \eqref{eq:kappa_pose} are used to derive
\begin{align}
	 \kappa_x = -r_{31}r_{22} / p_z
  \quad\text{and}\quad
  \kappa_y = -r_{31}r_{11} / p_z.
  \label{eq:kappa_xy_pose}
\end{align}
Note that a sanity check using \eqref{eq:kappa_xy_pose} shows that \eqref{eq:kappa_pose} can be recovered up to the sign.
It results in $\kappa = \operatorname{abs}\left(r_{31}/p_z\right)$, which is equivalent to \eqref{eq:kappa_pose} considering the used assumptions, \textit{i.e.}, $p_z > 0$ and $\kappa \geq 0$, which implies $-r_{31} = \sin\left(\kappa l\right) \geq 0$.

For a given arc space parameter choice, one uses either \eqref{eq:theta_orientation} and \eqref{eq:kappa_pose}, or \eqref{eq:kappa_xy_pose} to find the right set of parameters.
Note that, for a given tip pose as a combination of \eqref{eq:R_measured} and \eqref{eq:p_measured}, one can trivially combine the solutions from both previous cases.
However, the goal is to derive alternative solutions with potentially simpler expressions.

\paragraph{Recovering Full Pose from Tip Position}

From the previous derivation, it is clear that the tip position provides more information.
One intermediate question arises; Can the tip orientation be reconstructed using tip position?

To reconstruct the rotation matrix \eqref{eq:R_indep}, expressing the trigonometric functions $\cos\left(\theta\right)$, $\sin\left(\theta\right)$, $\cos\left(\kappa l\right)$, and $\sin\left(\kappa l\right)$ in terms of $\boldsymbol{p}$ is necessary.
For the derivation, \eqref{eq:theta_position} is used and after few steps, this leads to
\begin{align}
    \cos\left(\theta\right) = p_y / \sqrt{p_x^2 + p_y^2} 
    \quad\text{and}\quad 
    \sin\left(\theta\right) = p_x / \sqrt{p_x^2 + p_y^2}.\nonumber
\end{align}
Alternatively, \eqref{eq:kappa_xy_position} and \eqref{eq:kappa_position} can be combined.
For the other set, \eqref{eq:kappa_position} and \eqref{eq:p_z_position} are combined leading to
\begin{align}
    \sin\left(\kappa l\right) &= p_z\sqrt{p_x^2 + p_y^2}\, /\, \left( p_x^2 + p_y^2 + p_z^2 \right),\quad\text{whereas}\nonumber\\
    \cos\left(\kappa l\right) &= p_z^2\, /\, \left( p_x^2 + p_y^2 + p_z^2 \right) \nonumber
\end{align}
is derived involving the sum of squares of \eqref{eq:p_x_position} and \eqref{eq:p_y_position}.
Using this set of equations, $\boldsymbol{R}$ and, therefore, the whole pose can be reconstructed by only measuring $\boldsymbol{p}$.
\section{Unsuitable Representation as Pitfall}
\label{appendix:unsuitable_representations}

We share our insights on potential pitfalls arising from unsuitable joint and arc space representations.

\subsection{Absolute actuation length as representation}
\label{appendix:absolute_actuation_length}

Especially for TDCR, the actuation length \eqref{eq:tendon_length_literature} is a common choice for a joint representation in the literature.
However, we argue that this representation can obstruct the underlying structure of the joint space, which hinders a mapping to the embedded manifold.
We present three arguments to advocate in favour of displacement over actuation length as representation.

First, using actuation length, the displacement constraint \eqref{eq:sum_rho} does not equate to zero, and this non-zero component is dependent on $n$.
Therefore, the constraint appears to be in a non-general form, and one might think $n$ approaches are needed, which is akin to solving a quadratic function that is not presented in a normal form, \textit{i.e.}, $x^2 + px + q = 0$.

Second, as the analogy in Appendix~\ref{appendix:kirchhoff} suggests, the displacement constraint might not be a geometric constraint but rather a non-holonomic constraint.
Therefore, the constant value will vanish.

Third, a displacement-actuated joint for continuum robots is akin to a prismatic joint for ridged serial-kinematic robots.
A prismatic joint value is a relative and not an absolute value, avoiding unnecessary caring of a constant shift, making derived approaches more complex.
In fact, the property \eqref{eq:properties_vanishing_bias} eliminates a bias term.

\subsection{Various arc space representations}
\label{appendix:arc_space_representation}

We review existing representations of the arc space (see Table~\ref{tab:arc_space_representation}) for constant segment length $l > 0$.
The most commonly used representation is given by
\begin{align}
    \arg \left\{\text{arc space}\right\} \rightarrow \left(\kappa, \theta\right)
    \nonumber
\end{align}
according to the review by \cite{WebsterJones_IJRR_2010}.
This representation can be simply called \textit{curvature-angle representation}.
To avoid double coverage, $\kappa \in \mathbb{R}_{\geq 0}$ and $\theta \in \left[0, 2\pi\right)$ or, alternatively, $\kappa \in \mathbb{R}$ and $\theta \in \left[0, \pi\right)$.

\begin{table}[t]
    \renewcommand*{\arraystretch}{1.4}
    \caption{
    Arc space representation in a nutshell.
    }
    \label{tab:arc_space_representation}
    \centering
    \begin{tabular}{r r} 
        \toprule
        \multicolumn{1}{N}{$\arg \left\{\text{arc space}\right\}$} & \multicolumn{1}{N}{naming} \\
		\cmidrule(r){1-1}
		\cmidrule(l){2-2}
        $\left(\kappa, \theta\right)$ & curvature-angle representation \\[.25em]
        $\left(\kappa_x, \kappa_y\right)$ & curvature-curvature representation \\[.25em]
        $\left(\phi, \theta\right)$ & angle-angle representation \\[.25em]
        $\left(\kappa\cos\left(\theta\right), \kappa\sin\left(\theta\right)\right)$ & non-linear combination \\[.25em]
        \bottomrule
    \end{tabular}
\end{table}

Another common representation utilizes
\begin{align}
    \arg \left\{\text{arc space}\right\} \rightarrow \left(\kappa_x, \kappa_y\right),
    \nonumber
\end{align}
where $\kappa_x \in \mathbb{R}$ and $\kappa_y \in \mathbb{R}$ are two curvatures, which are called Cartesian curvature components by \cite{DupontRucker_et_al_JPROC_2022}.
This representation can be called \textit{curvature-curvature representation}.

For completeness, another representation utilizes $\phi = \kappa l$, which makes this representation equal to the curvature-angle representation.
Following the naming convention, \textit{angle-angle representation} is defined by
\begin{align}
    \arg \left\{\text{arc space}\right\} \rightarrow \left(\phi, \theta\right).
    \nonumber
\end{align}
Both angles can be restricted to $\phi \in \left[0, 2\pi\right)$ and $\theta \in \left[0, 2\pi\right)$.
Furthermore, both angles relate to azimuth and elevation angles \citep{DupontRucker_et_al_JPROC_2022}.

In recent years, a non-linear combination of the curvature $\kappa$ and bending angle $\theta$ has been used more often.
We denote
\begin{align}
    \arg \left\{\text{arc space}\right\} \rightarrow \left(\kappa\cos\left(\theta\right), \kappa\sin\left(\theta\right)\right),
    \nonumber
\end{align}
which is another possibility to represent the arc space.

Commonly the curvature $\kappa_x$ and $\kappa_y$ are transformed into $\kappa = \sqrt{\kappa_x^2 + \kappa_y^2}$ and $\theta = \arctantwo\left(\kappa_y, \kappa_x\right)$.
Both can be used to map from the curvature-curvature representation to the curvature-angle representation.
To reverse the map, let us apply trigonometric functions to $\theta$, \textit{i.e.}, $\cos(\theta)$ and $\sin(\theta)$, to derive $\cos(\theta) = \kappa_x / \sqrt{\kappa_x^2 + \kappa_y^2}$ and $\sin(\theta) = \kappa_y / \sqrt{\kappa_x^2 + \kappa_y^2}$.
Further simplification leads to $\kappa_x  = \kappa\cos(\theta)$ and $\kappa_y  = \kappa\sin(\theta)$.
This set of equations can be used to map from the curvature-angle representation to the curvature-curvature representation.
Two remarks should be noted.
First, for constant segment length $l$, curvature-angle representation and angle-angle representation are linear dependent due to $\phi = \kappa l$.
Second, the representation that combines the curvature and angle in a non-linear way is the curvature-curvature representation in disguise.

In recent works, the angle-angle representation has been identified as problematic.
For instance, \cite{DellaSantinaBicchiRus_RAL_2020} demonstrates pathological behaviors can arise when controlling a soft robot represented by the angle-angle representation.
Note that the linear relation $\phi = \kappa l$ indicates that popular curvature-angle representation has the same disadvantages as the angle-angle representation.
On the other hand, using the curvature-curvature representation avoids coordinate singularities \citep{DupontRucker_et_al_JPROC_2022} and allows to formulate compact kinematic expressions \citep{AllenAlbert_et_al_RoboSoft_2020}.
To conclude, depending on the choice of arc space representation, a derived approach can behave pathologically and can be overly complicated in its formulation.

\bibliographystyle{asrl}
\bibliography{references}

\end{document}